\documentclass[10pt,twocolumn,letterpaper]{article}

\usepackage{wacv}              % To produce the CAMERA-READY version
\usepackage{graphicx}
\usepackage{amsmath}
\usepackage{amssymb}
\usepackage{booktabs}
\usepackage[table]{xcolor}

\usepackage[accsupp]{axessibility}

\usepackage[pagebackref,breaklinks,colorlinks]{hyperref}

\usepackage[capitalize]{cleveref}
\crefname{section}{Sec.}{Secs.}
\Crefname{section}{Section}{Sections}
\Crefname{table}{Table}{Tables}
\crefname{table}{Tab.}{Tabs.}

\newcommand{\rev}{}

\usepackage{booktabs, multirow} % for borders and merged 
\usepackage{soul}% for underlines
\usepackage{subcaption}
\usepackage{tabularx}
\usepackage{listings}
\usepackage{tabularx}

\usepackage{changepage,threeparttable} % for wide tables
\usepackage{booktabs, multirow} % for borders and merged 

\usepackage{enumitem}

\definecolor{highlight-blue}{HTML}{388AF5}
\definecolor{LLGray}{gray}{0.91}
\definecolor{LGray}{gray}{0.94}

\definecolor{highlight-blue}{HTML}{388AF5}
\definecolor{LLGray}{gray}{0.91}
\definecolor{LGray}{gray}{0.94}
\definecolor{Green}{RGB}{4, 130, 25}

\begin{document}

%%%%%%%%% TITLE - PLEASE UPDATE
% \title{Atypicality-Aware Verbalization Helps Reasoning about Visual Advertisements} 
\title{Benchmarking VLMs' Reasoning About Persuasive Atypical Images}
% Suggestions:
% Atypicality Insights: Reasoning on Persuasive Visual Media
% Unveiling Atypicality: Evaluating Vision Language Models in Persuasive Visual Media
% Understanding Persuasive Visual Media: Atypicality and Reasoning in Vision Language Models

\author{Sina Malakouti\textsuperscript{\rm 1,*}
\hspace{0.25cm}
Aysan Aghazadeh\textsuperscript{\rm 1,*}
\hspace{0.25cm}
Ashmit Khandelwal\textsuperscript{\rm 2}
\hspace{0.25cm}
Adriana Kovashka\textsuperscript{\rm 1}\\
\textsuperscript{1}University of Pittsburgh \hspace{1cm}
\textsuperscript{2}BITS Pilani \\
{\tt\small \{sem238, aya34\}@pitt.edu \hspace{0.25cm}
f20200980@goa.bits-pilani.ac.in 
\hspace{0.25cm}
kovashka@cs.pitt.edu}
}

% \author{Sina Malakouti\textsuperscript{\rm *}\\
% University of Pittsburgh\\
% {\tt\small sem238@pitt.edu}
% % For a paper whose authors are all at the same institution,
% % omit the following lines up until the closing ``}''.
% % Additional authors and addresses can be added with ``\and'',
% % just like the second author.
% % To save space, use either the email address or home page, not both
% \and
% Aysan Aghazadeh\textsuperscript{\rm *}\\
% University of Pittsburgh\\
% {\tt\small aya34@pitt.edu}
% \and
% Ashmit  Khandelwal\\
% BITS Pilani\\
% {\tt\small f20200980@goa.bits-pilani.ac.in}
% \and
% Adriana Kovashka\\
% University of Pittsburgh\\
% {\tt\small kovashka@cs.pitt.edu}
% \and
% }
\maketitle

\renewcommand*{\thefootnote}{\fnsymbol{footnote}} % Use symbols instead of numbers for footnotes
\footnotetext[1]{Equal Contribution. Listing order is random.}

%%%%%%%%% ABSTRACT
% \begin{abstract}
%    The ABSTRACT is to be in fully justified italicized text, at the top of the left-hand column, below the author and affiliation information.
%    Use the word ``Abstract'' as the title, in 12-point Times, boldface type, centered relative to the column, initially capitalized.
%    The abstract is to be in 10-point, single-spaced type.
%    Leave two blank lines after the Abstract, then begin the main text.
%    Look at previous WACV abstracts to get a feel for style and length.
% \end{abstract}

\begin{abstract}
    
% Vision language models (VLM) have rapidly progressed, especially with integrating large language models (LLMs), exhibiting strong zero-shot generalization across diverse tasks.
Vision-language models (VLMs) have shown strong zero-shot generalization across various tasks, especially when integrated with large language models (LLMs). 
However, their ability to comprehend rhetorical and persuasive visual media, such as advertisements, remains understudied. Ads often employ \textbf{atypical imagery}, using surprising object juxtapositions to convey shared properties. For example, Fig.~\ref{fig:intro_examples} (e) shows a beer with a feather-like texture. This requires \textbf{advanced reasoning} to deduce that this atypical representation signifies the beer's lightness.

We introduce three novel tasks, \rev{Multi-label Atypicality Classification, Atypicality Statement Retrieval, and Atypical Object Recognition,} to benchmark VLMs' understanding of atypicality in persuasive images. We evaluate how well VLMs use atypicality to infer an ad's message and test their reasoning abilities by employing
%on the action-reason retrieval (ARR) task and develop 
semantically challenging negatives. Finally, we pioneer atypicality-aware verbalization by extracting comprehensive image descriptions sensitive to atypical elements.

Findings reveal that: (1) VLMs lack advanced reasoning capabilities compared to LLMs; 
% (2) LLMs exhibit superior reasoning compared to VLMs;
(2) simple, effective strategies can extract atypicality-aware information, leading to comprehensive image verbalization; (3) atypicality aids persuasive ad understanding. Code and data is available at \href{https://aysanaghazadeh.github.io/PersuasiveAdVLMBenchmark/}{aysanaghazadeh.github.io/PersuasiveAdVLMBenchmark/}
% These results underscore the importance of atypicality in interpreting persuasive visual media.
% Code and data will be made available. 

\end{abstract}

\section{Introduction}
\label{sec:intro}
% Paragraph1 : Define/explain advertisement, atypicality and their challenges

In visual media, particularly advertisements, creators employ \emph{creative} visual rhetoric to capture attention and convey memorable, powerful messages. They intentionally deviate from realism, depicting objects in unique and atypical ways \cite{reinartz2013creativity,mcquarrie1999visual}. Creative ads that are ``out of the ordinary'' or ``connect objects that are usually unrelated'' can generate twice as much revenue as non-creative ads \cite{reinartz2013creativity}. 

\begin{figure}[t]
        \includegraphics[width=\linewidth]{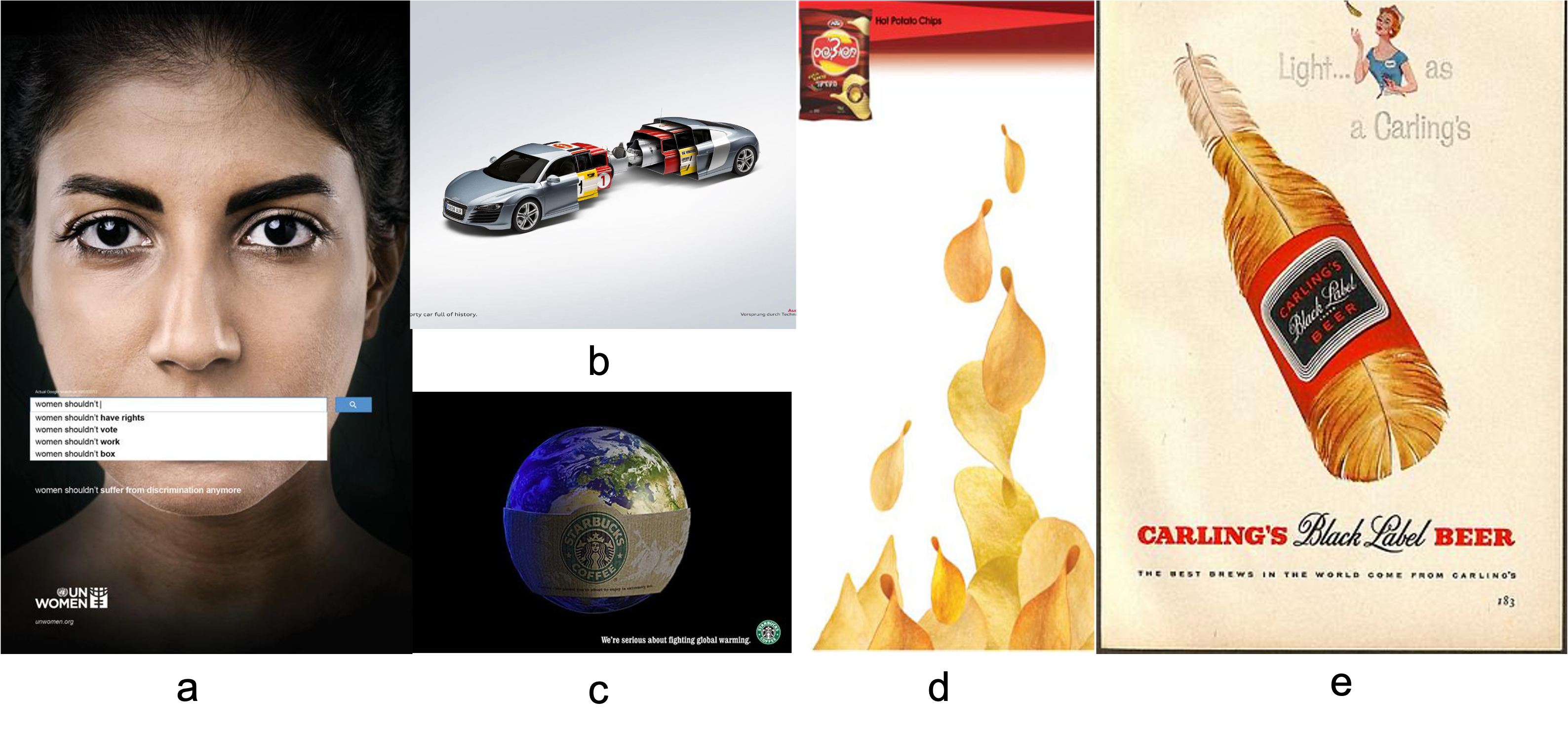}
        % \caption{}
        % \label{fig:intro_a}
    
    \caption{\textbf{Atypicality categories.} We study four types of atypicality from  \cite{yeinterpreting}: Texture Replacement 1, Texture Replacement 2,  Object Inside Objects, Object Replacement (defined in Sec.~\ref{sec:atyp-tasks}).}
    \label{fig:intro_examples}

\end{figure}

\textit{Atypical imagery} in ads often involves transforming objects metaphorically \cite{williamson1978decoding,yeinterpreting}.
These creative transformations are not random; they are carefully chosen to convey specific ideas ~\cite{williamson1978decoding}.
For example, Fig.~\ref{fig:intro_examples}(a) depicts a text box as tape to suggest silencing, while in (d), potato chips are shown as flames to metaphorically represent spiciness, borrowing properties from fire (hotness symbolizing flavor).
Understanding these atypical images requires more than just recognizing objects. It requires advanced reasoning skills, including knowledge of cultural contexts and social norms, posing a significant challenge for AI systems.

% Atypicality often involves diverse and creative metaphorical object transformations\cite{williamson1978decoding,yeinterpreting}. Understanding these atypical objects is essential to decode the message of ads~\cite{williamson1978decoding}. For example, in Fig.~\ref{fig:intro_examples}(a), a text box is depicted as tape to signify silencing, while in (d), potato chips are shown as flames to metaphorically represent spiciness, borrowing properties from fire (hotness symbolizing flavor). 
% Interpreting atypicality and their connection to the ad message requires advanced reasoning abilities incorporating background knowledge and social/cultural understanding.

% Paragraph 2: explain VLMs, evaluations, etc. 
% Despite significant advancements in vision-language models (VLMs) and their applications in various tasks such as image captioning and visual question answering, there remains a substantial gap in understanding complex, rhetorical visual media like persuasive advertisements. Previous benchmarks, such as COCO and VQA, focus primarily on literal image understanding and fail to capture the nuanced reasoning required for interpreting atypical imagery often found in advertisements. Our work introduces three novel tasks specifically designed to address this gap: multi-label atypicality classification, atypicality statement retrieval, and atypical object recognition. These tasks not only challenge the models to understand atypical visual elements but also to reason about their implications in the context of persuasive messages, thereby representing a significant departure from existing benchmarks.

Modern pretrained vision-language models (VLMs) like LLaVA \cite{liu2023visual,liu2023improvedllava} demonstrate strong visual understanding across various tasks such as recognition \cite{NEURIPS2023_761c3284}, and capabilities like zero-shot generalizability \cite{safaei2024active,ge2023improving}. However, there is a lack of in-depth study on VLMs’ ability to understand complex persuasive images such as advertisements.
% which rely on creative design techniques like atypical imagery. 

We address this gap by introducing three novel tasks over PittAds \cite{hussain2017automatic} to evaluate VLMs' understanding of atypicality: (1) multi-label atypicality classification, where the model predicts the type of atypicalities in the image; (2) atypicality statement retrieval, \rev{where the model retrieves correct atypicality statements describing the atypicality relation among objects}; (3) atypical object recognition, \rev{where the model generates objects to complete an atypicality statement based on a given relation}. These tasks are essential as prior works' binary classification oversimplifies atypicality's nuanced nature. Our evaluation shows that although VLMs struggle with direct atypicality inference, they can extract valuable information about atypical aspects.

Next, we investigate how atypicality influences understanding an ad's message. We use the action-reason retrieval (ARR) task \cite{hussain2017automatic,yeinterpreting}, which requires models to identify the suggested action (e.g., ``buy these chips") and its rationale (e.g., ``because they are spicy"). However, to rigorously test the model's reasoning, we introduce semantically challenging negative options rather than mining hard negatives from other images \cite{ye2018advise,jia-etal-2023-kafa}. For example, we generate statements that include wrong action (e.g., ``don't buy these chips") or wrong rationale (e.g., ``because they are sweet"). This prevents VLMs from ruling out negatives by merely comparing objects in the image and options. Our evaluation shows a significant performance drop when VLM is faced with hard negatives (e.g., LLaVA drops by 67.51\%). 

Finally, we hypothesize that \textit{deep atypicality understanding enhances action-reason retrieval performance}. To test this, we propose an atypicality-aware verbalization.
 % method using LLaVA 
% \cite{liu2023visual}
Using simple prompting strategies, we generate a comprehensive atypicality-sensitive ad verbalization to predict the corresponding atypicality statement.
% This statement is then combined with the atypicality-aware verbalization and fed into an LLM classifier, which effectively combines the benefits of visual understanding and reasoning to predict the final action-reason.
Then, an LLM integrates this statement with atypicality-aware verbalization to retrieve the final action-reason, effectively combining the benefit of both visual understanding and reasoning.

Our proposed framework achieves state-of-the-art performance on the ARR task. Interestingly, when a VLM is given both the image and our atypicality-aware verbalization, its performance on the ARR task declines (e.g., LLaVA($I + \mathcal{T}_\mathcal{V}$) shows a 1.71 point drop in performance compared to LLaVA($I$)) and it is significantly underperformed compared to LLM. This stark contrast highlights a critical gap: \textit{VLMs lack the advanced reasoning capabilities of LLMs when interpreting complex, atypical visual media}. To summarize, our contributions are: 
\begin{enumerate}[nolistsep,noitemsep]
    \item We introduce three novel tasks for understanding atypicality in persuasive media. 
    \item We pioneer the use of atypicality inference in action-reason retrieval and are the first to benchmark VLMs/LLMs for advertisement understanding. 
    \item We generate semantically challenging negatives using GPT-4 for action-reason retrieval, revealing VLMs' reasoning limitations in interpreting atypical ads.
\end{enumerate}
We hope this work inspires the inclusion of persuasive ads in VLM benchmarks, fosters the development of robust models for complex visual media, and offers insights for creating more effective advertisements.

\section{Related Works}
\label{sec:related_works}

\textbf{Creativity in advertising} has long been of interest in advertising research. It has been broken down into categories, and its impact on the effectiveness of ads has been measured. Both \cite{reinartz2013creativity} and \cite{smith2007modeling} define the categories as originality, flexibility, synthesis, elaboration, and artistic value, which capture different shades of divergence from the ordinary. \emph{Atypicality} most directly maps to \emph{synthesis}. However, these creativity strategies have \textbf{not been explored in computer vision} for predicting the message of an ad. 

\textbf{Advertisement image understanding.}
The PittAds Dataset~\cite{hussain2017automatic} introduced the action-reason retrieval task, establishing a baseline for automatic ad understanding. However, most studies have not explicitly captured advertising-specific strategies for this task, nor have they addressed atypicality. \cite{ye2018advise} incorporated symbolism, but the gains were minimal. Others utilized scene-text \cite{dey2021beyond,kalra2020understanding}, graph-based methods to incorporate external knowledge \cite{Ye_2021_WACV}, and CLIP \cite{radford2021learning} %and a feature adaptation strategy 
for brand name understanding \cite{jia-etal-2023-kafa}. 
\cite{bhattacharya2023video} used Automatic Speech Recognition, OCR, WikiData and BLIP-2 \cite{li2023blip} to describe the stories of video ads.
\cite{akula2023metaclue} analyzed metaphors in ads.
Yet, the impact of atypicality on ad image understanding remains unexplored.
The only exception is \cite{guo2021detecting}, which proposed a self-supervised approach to classify images as typical or atypical but did not classify the type of atypicality nor use them for action-reason prediction.
%Inferring atypicality requires a deep understanding of the image and reasoning ability, which poses a challenge for conventional CV models and discriminative VLMs (e.g., CLIP \cite{radford2021learning}). In this work, we leverage generative VLMs (e.g., LLaVA \cite{liu2023improvedllava}) for atypicality-aware verbalization of images.  our study focuses on the impact of atypicality in ad image understanding. We introduce 3 novel tasks to evaluate VLMs' ability to understand atypicality, highlight the lack of reasoning ability of VLMs, and propose to produce atypicality-aware verbalizations and detect atypicality statements, significantly improving action-reason retrieval performance. Our work diverges by targeting atypicality and action-reason retrieval, using VLMs for verbalization and extensive evaluation of VLM/LLM reasoning abilities.  

\textbf{Vision-language models.} 
We benchmark pre-trained VLMs and LLMs on tasks involving atypicality and advertisement image understanding, focusing on their zero-shot reasoning capabilities. We use pre-trained VLMs to verbalize advertisement images for LLMs. 
Given the substantial computational power required for training and fine-tuning large models, off-the-shelf, frozen models are typically used \cite{tsimpoukelli2021multimodal,alayrac2022flamingo}. 
Techniques to align visual and textual features without parameter updates include optimizing image encoders \cite{tsimpoukelli2021multimodal}, inserting cross-attention layers \cite{alayrac2022flamingo}, prompt learning \cite{zhou2022learning,zhou2022conditional, chen2023understanding,khattak2023maple}, and employing external transformers \cite{mokady2021clipcap,li2023blip,dai2023instructblip,liu2023improvedllava}. However, direct application of models like \cite{li2023blip,radford2021learning} may miss hidden messages by focusing more on visual context than semantics. Note that we restrict our experiments to the zero-shot inference setting.
%, with plans to integrate learned prompts in future work further to enhance our atypicality-aware verbalization approach, similar to \cite{yao2023visual}.

% \noindent 
\textbf{Language models for multi-modal reasoning.}
Recent studies \cite{dziri2024faith, zhang2023multimodal,bhattacharya2023video,li-etal-2024-enhancing-advanced} have explored LLMs for reasoning tasks, including chain-of-thought reasoning \cite{wei2022chain}.
%which requires generating step-by-step rationales. 
Some works leverage LLMs in multi-modal reasoning. \cite{zhang2023multimodal,wang2024t,zheng2023ddcot} extended chain-of-thought to a multimodal context.\cite{lan2023improving} uses an image-captioning model followed by `reasoning questions' to aid an LLM in answering the main question. Related/concurrent works like \cite{you2023idealgpt} improve zero-shot reasoning by iteratively asking and answering questions with 3 VLM/LLM, and \cite{mitra2024compositional} uses scene-graphs to enhance compositional reasoning.  \cite{lu2024chameleon,zhang2023multimodal} devised sophisticated LLM-augmented tools for task subdivision and external tool selection. In contrast, this paper challenges the reasoning ability of VLMs on complex persuasive images through novel atypicality tasks and action-reason retrieval. It %highlights their limitations compared to LLMs using semantically hard negatives and significantly 
improves performance with a more lightweight atypicality-aware verbalization, and no external tools are needed. 

\begin{figure*}[!tp]
    \centering
    \includegraphics[width=1\textwidth]{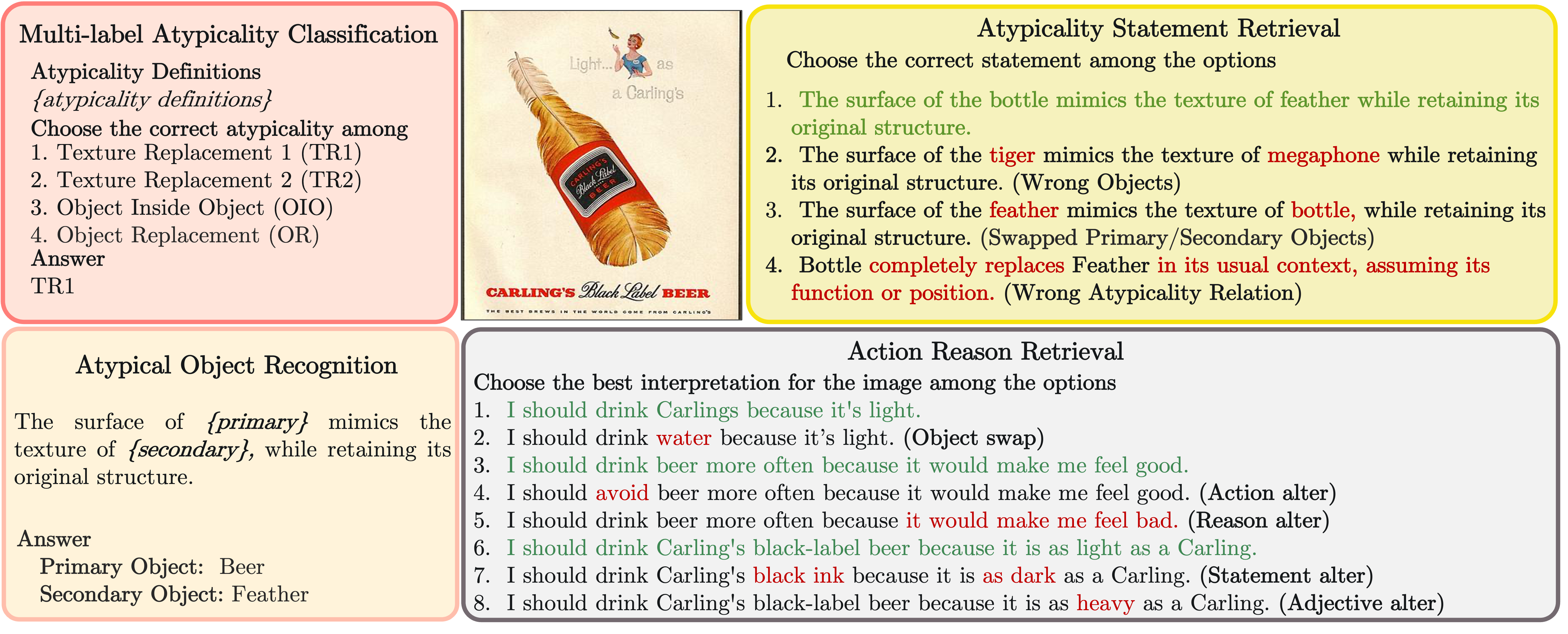}
    \caption{
\textbf{Atypicality Understanding and Action-Reason Retrieval Tasks}. We introduce three tasks: Multi-label Atypicality Classification, Atypicality Statement Retrieval, and Atypical Object Retrieval. Incorrect/correct phrases/statements are in \textcolor{red}{red}/\textcolor{green}{green}.}
    \label{fig:tasks}
\end{figure*}

\textbf{VLM evaluation}. Several vision-language benchmarks focus on tasks like recognition \cite{goyal2017making}, captioning \cite{chen2015microsoft}, commonsense reasoning \cite{vedantam2015learning,bitton2023vasr}, VCR \cite{zellers2019recognition}, and compositional reasoning \cite{thrush2022winoground}. 
However, these benchmarks typically involve simple scenes that fail to test the model's reasoning abilities beyond the basic interpretation of literal imagery.
% straightforward scenes that do not challenge the model's reasoning abilities beyond literal imagery. 
% Recently, ROME~\cite{zhou2023rome} generated counter-intuitive images focusing on 5 primitive common sense types (e.g., color and size) to challenge models' ability in object, attribute, and spatial relation recognition. Similarly, WHOOPS! \cite{whoops!} addressed this by creating 500 synthetic scenes by placing `normal' objects in unusual contexts (e.g., a snowplow in a desert and Einstein holding a smartphone).
Recently, ROME~\cite{zhou2023rome} generated counter-intuitive images focusing on 5 primitive common sense types (e.g., color and size) to challenge models' ability in object, attribute, and spatial relation recognition. Similarly, WHOOPS! \cite{whoops!} addressed this by creating 500 synthetic scenes by placing `normal' objects in unusual contexts (e.g., a snowplow in a desert and Einstein holding a smartphone).
Atypical advertising images differ from those in WHOOPS! and ROME in \textbf{two important ways}: First, they are real ads created by human designers to intentionally convey a particular message (e.g., cigar replacing a bullet, to highlight the dangers of smoking), rather than merely aiming to be unusual. 
% This requires models to detect atypicality and atypical objects and reason about their impact on the ad's meaning, making ad image understanding a more realistic and challenging benchmark for evaluating VLMs' reasoning abilities.
This requires models to detect atypicality and atypical objects and reason about their impact on the ad's message. Ad image understanding becomes a more realistic and challenging benchmark for evaluating VLMs' reasoning ability.
% Second, atypical ads feature more categories of atypicality than placing an object out of context or altering primitive attributes, as WHOOPS! and ROME do, respectively.
Second, atypical ads involve more types of atypicality than simply placing an object out of context, as in WHOOPS!, or altering primitive attributes, as in ROME.

\section{Methodology}
\label{sec:method}

%Ads often feature objects that combine two elements to showcase shared properties. Recognizing these objects and deciphering their unique relationships is crucial for interpreting visual media designs. However, these objects are unusual and out-of-distribution, making their recognition challenging. Identifying the shared property and its connection to the ad's message requires advanced multi-step reasoning. 

This paper evaluates VLM/LLM understanding of atypical advertisements. We address two key questions: (1) Are current VLMs capable of reasoning about atypicality and understanding advertisements? (2) What is the impact of atypicality on understanding ad images? 

Unlike prior works \cite{guo2021detecting} that only classify images as typical or atypical, we propose three new tasks: Multi-label Atypicality Classification (MAC), Atypicality Statement Retrieval (ASR), and Atypical Object Recognition (AOR). MAC predicts multiple categories of atypicality in the image. ASR uses additional annotations to identify objects involved in the atypical portrayal (e.g., "The surface of the \underline{bottle} mimics the texture of \underline{feather}"). AOR evaluates VLMs' visual reasoning by identifying primary and secondary objects in atypical relation.

Our analysis shows while VLMs initially struggle with MAC and ASR tasks, they can extract valuable insights about atypical aspects of images. Leveraging these insights, we develop an atypicality-aware image verbalization. To detect atypicality, we use the prompt \textit{UH: What is unusual about this image?}. Atypicality adds depth to the content of the advertisement, complementing surface-level content like objects and scene descriptions. Thus, we combine $UH$ and surface-level content to construct the final verbalization.
It is then passed to an LLM for the action-reason inference task. We elaborate on the tasks and pipeline below.

\begin{figure*}[!tp]
    \centering
    \includegraphics[width=1\textwidth]{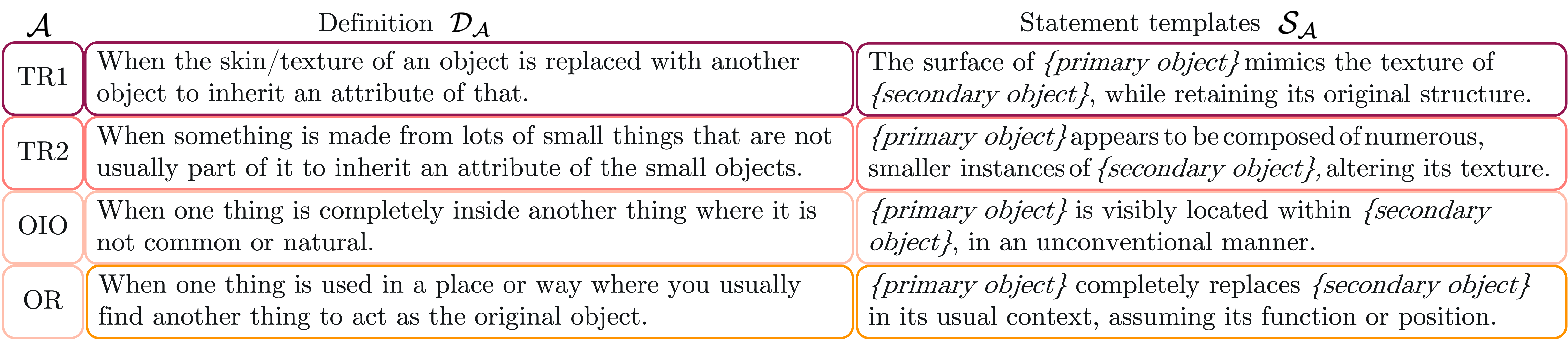}
    \caption{Atypicality definitions and atypicality relation statements.}
    \label{fig:atyp_def_stmt}
\end{figure*}

\subsection{Proposed Atypicality Understanding Tasks}
\label{sec:atyp-tasks}

Ye et al. \cite{yeinterpreting} devised a taxonomy of atypicality based on object transformations. In this work, we focus on the subset of atypicality categories that entail two objects (examples in Fig~\ref{fig:intro_examples}): (1) Texture Replacement 1 (TR1): Objects’ texture borrowed from another object, (2) Texture Replacement 2 (TR2): Texture created by combining several small objects, Object Inside Object (OIO):  An object is completely or partially inside of another object, and (4) Object Replacement (OR): The whole object appearing in the context normally associated with another.
We define the following new atypicality understanding tasks, shown in Fig.~\ref{fig:tasks}. 

\textbf{Multi-label Atypicality Classification (MAC).} 
Unlike prior works \cite{guo2021detecting} that only detect the presence of atypicality, we formulate atypicality detection as a multi-label classification task. The PittAds dataset provides three annotations of atypicality per image from different annotators, which may vary by type. For example, Fig.\ref{fig:intro_examples}(c) can be classified as `Object Inside Object' (Earth inside a cup sleeve) and `Object Replacement' (Earth replaces coffee cup). \rev{Some annotators may even label it as `Not Atypical' (NA), creating five possible labels.} %suggesting the image is typical.
MAC challenges VLMs to predict the relevant atypicality categories for an image based on atypicality definitions denoted as $\mathcal{D}_\mathcal{A}$ (prompts in supp). \rev{Due to the complexity of differentiating between these categories, we extend the definitions provided by \cite{yeinterpreting} as shown in Fig.~\ref{fig:atyp_def_stmt}}: they not only distinguish different atypicality categories but also hint at how atypicality impacts the image's interpretation (e.g., Fig.~\ref{fig:intro_examples} is TR1, where the beer's texture is replaced by a feather to advertise its lightness). 
% $a \in \mathcal{A}$ %for each image $I$

\begin{figure*}[t]
    \centering
    % First image (full width)
    \begin{subfigure}{0.96\textwidth}
    \centering
        \includegraphics[width=\textwidth]{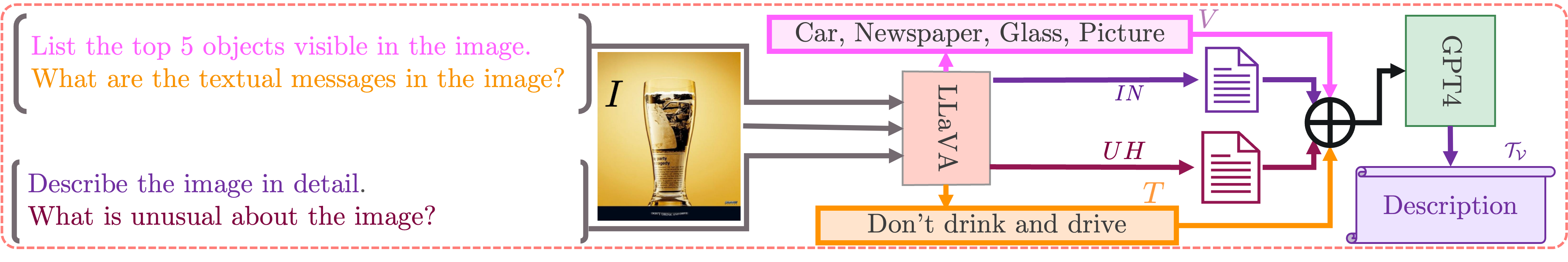}
        \caption{Image Verbalization}
        \label{fig:imagedescription}
    \end{subfigure}
    \\
    \begin{subfigure}{0.48\textwidth}
        \centering % Centering the image inside the subfigure
        \includegraphics[width=\textwidth]{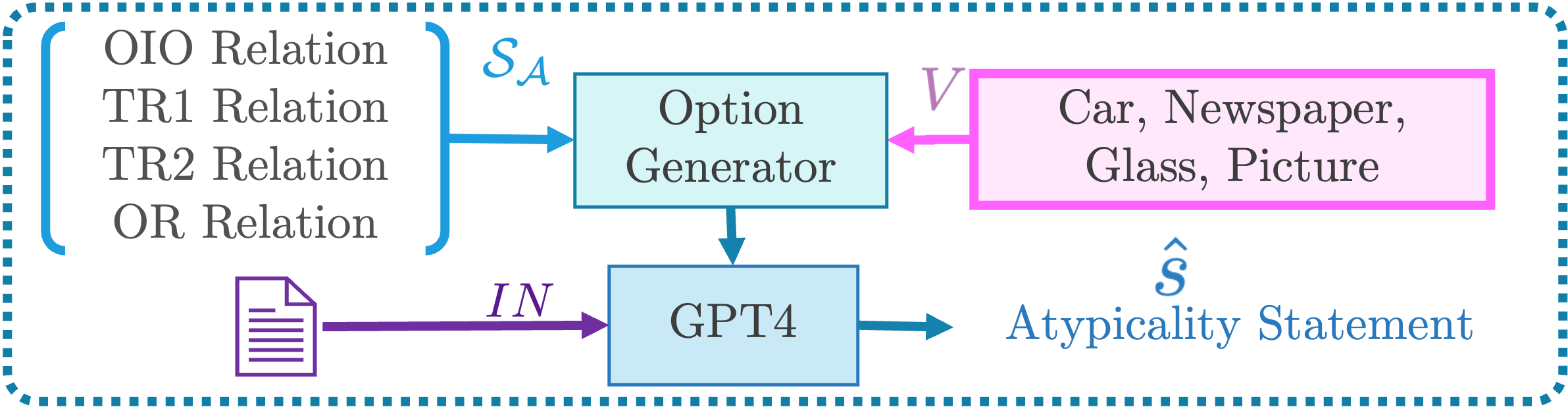}
        \caption{Atypicality Statement Detection}
        \label{fig:method_asd}
    \end{subfigure}
    % \hfill % Space between the second and third images
    % Third image (half width, aligned with the second image)
    % Second image (half width, aligned with the third image)
    \begin{subfigure}{0.48\textwidth}
    
        \centering % Centering the image inside the subfigure
        \includegraphics[width=\textwidth]{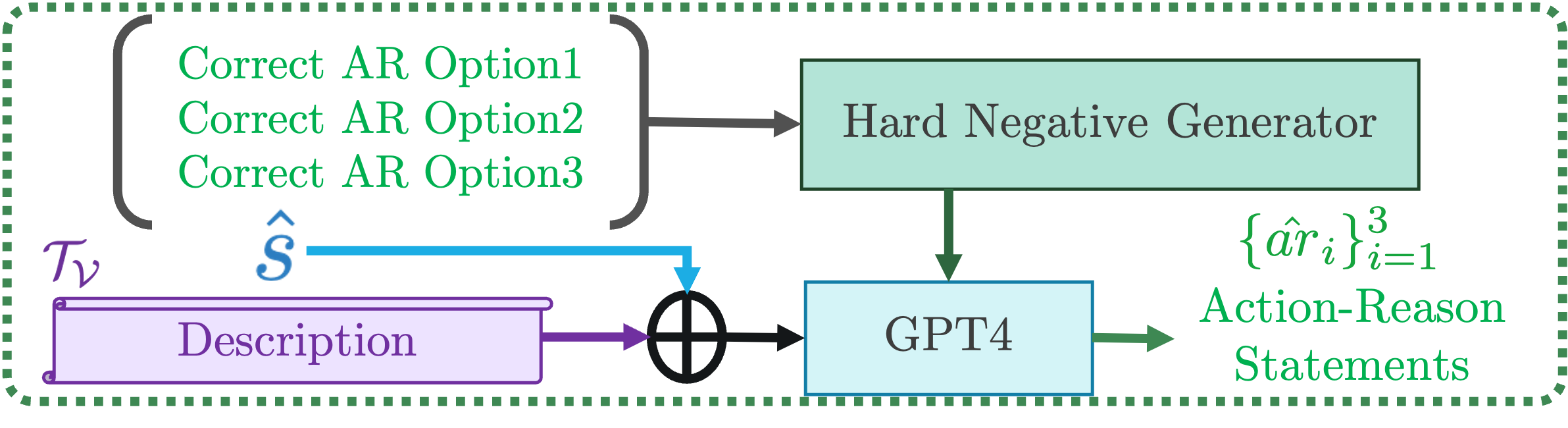}
        \caption{Action-Reason Retrieval}
        \label{fig:method_actionreason}
    \end{subfigure}
    
    \caption{Our approach consists of three steps: (a) \textbf{Image verbalization}: We first prompt LLaVA to obtain top-5 objects ($V$), scene-text ($T$), scene description $IN$, and unusualness $UH$. Then we combine all the information to obtain a uniform description $\mathcal{T}_{\mathcal{V}}$. (b) \textbf{Atypicality Statement Detection}: We utilize $V$ and atypicality statement templates $\mathcal{S}_\mathcal{A}$ to generate the options which are then used along with $IN$ to retrieve the atypicality statement $\hat{s}$. (c) \textbf{Action-Reason Retrieval}: We input $\hat{s}$ along with verbalization $\mathcal{T}_\mathcal{V}$ to retrieve action-reason. 
    % (a) We prompt LLaVA to verbalize the image. we combine list of objects $V$, textual slogans $T$,   \textit{ImageNarrator}} ($IN$), and \textit{UnusualHighligher}($UH$) using GPT-4. (b) Final verbalization $\mathcal{T}_\mathcal{V}$ is used to detect the atypicality statement $\hat{s}$ among options created by Option Generator by combining objects $V$, and statement templates $\mathcal{S}_\mathcal{A}$. (c) $\hat{s}$ is combined with $\mathcal{T}_\mathcal{V}$ for the final action-reason retrieval task.
    }
    \label{fig:method-overview}
% \vspace{-5pt}
\end{figure*}

% \noindent
\textbf{Atypicality Statement Retrieval (ASR).} ASR frames atypicality inference as retrieving a statement describing relations between two objects. Atypicality $\mathcal{A}$ is presented using templates $\mathcal{S_{\mathcal{A}}}$, as defined in Fig.~\ref{fig:atyp_def_stmt}.
Each statement $s = (a o^p, o^s)$ includes an atypicality type $a$, primary object $o^p$, and secondary object $o^s$ as described in Sec. 3.6 of \cite{yeinterpreting}. For TR1 and TR2, $o^p$ is the object with the new texture, and $o^s$ is the texture source. For OIO, $o^p$ is the object inside, and $o^s$ is the object outside. For OR, $o^p$ is the object placed in the context of another, and $o^s$ is the expected object. 
% The relationship $r$ is fixed based on the atypicality category.

Given an ad, ASR distinguishes the correct atypicality statement \rev{${s}^{+}=(a^+,o^{+p},o^{+s})$}, from \rev{a set of} negative statements $\{\Bar{s_{i}}\} \in {S}^{-}$. 
We generate negatives as follows: (1) \textbf{Wrong object}: replacing $o^{+p}$ and $o^{+s}$ with objects from $K$ random images, producing $2K$ negatives (e.g., $\Bar{s}_1 = (a^+, o^{p}_{1}, o^{s}_{1})$ and $\Bar{s}_2 = (a^+, o^{s}_{1}, o^{p}_{1})$ where $o^{p}_{1}/o^{s}_{1}$ are from a random image);
% We retrieve objects from $K$ random images in the dataset (i.e, $I_k=(a_k, o^{p}_{k}, o^{s}_{k})$ where $k = 1 $ to $K$) to create $2K$ negatives by replacing the $o^{+p}$ and $o^{+s}$ (e.g., $\Bar{s}_1 = (a^+, o^{p}_{1}, o^{s}_{1})$ and $\Bar{s}_2 = (a^+, o^{s}_{1}, o^{p}_{1})$); 
(2) \textbf{Wrong atypicality relation}: altering the relation with one not in the ground-truths (objects remain the same) to create up to 3 negatives.
% (i.e., $(\Bar{a}_i, o^{+p}, o^{+s})$).
% We keep original objects but alter the relation with one not in the ground-truth, to create up to 3 negatives (i.e., $(\Bar{a}_i, o^{+p}, o^{+s})$); and 
(3) \textbf{Swapped primary/secondary objects}: we create $(a^+,o^{+s},o^{+p})$. 
% to test the model's understanding of the relation between objects.
Thus, ASR tests the model's understanding of objects, their atypicality relation, and primary/secondary object roles. It bridges the gap between MAC and action-reason retrieval by detecting atypical statements and enhancing action-reason retrieval. We use $K=2$.
% ASR can be directly employed in the final method to detect the atypicality statement for enhancing action-reason retrieval, bridging the gap between MAC and action-reason retrieval. We select $K=2$.

% \noindent
\textbf{Atypical Object Recognition (AOR).}
To assess the VLMs' fine-grained visual understanding, we propose recognizing the primary and secondary objects in an atypical image. \rev{Given ad image and true atypicality},
% atypicality statement $s = (a^{+}, -, -)$ where true atypicality $a^{+}$ is only given
% % without the primary/secondary objects (i.e., $-$) is fed to 
% the model ($-$ means blank). 
\rev{the goal is to generate the correct primary/secondary objects to complete the atypicality statement.} AOR functions as a fill-in-the-blank task (i.e., generative) using using statement templates. 

\subsection{Proposed Approach} 
\label{sec:proposed-approach}
To explore the impact of atypicality in action-reason retrieval and compare the reasoning abilities of VLMs and LLMs, we propose an in-context learning method 
%We first verbalize the image based on insights from Sec.~\ref{sec:atyp-tasks} and adopt LLM classifier . 
% We believe that the current VLMs' limitations in commonsense reasoning hinder their ability to directly address atypicality detection and action-reason retrieval tasks. Therefore, we propose to use an LLM, which has shown strong reasoning ability for the final downstream task. However, since LLMs such as GPT-4 cannot process images directly, we use LLaVA to produce an effective verbalization $\mathcal{T}_\mathcal{V}$ that can be successfully used for atypicality understanding tasks and the subsequent action-reason retrieval. 
% The proposed method consists of three steps. (i) Atypicality-aware image verbalization: We use LLaVA \cite{liu2023visual} and LLM to generate an explicit and coherent verbalization $\mathcal{T}_{\mathcal{V}}$ that includes information about the atypicality in the image; (ii) Atypicality statement retrieval: We use LLM to retrieve the atypicality statement $\hat{s}=(\hat{a}, \hat{o}^p, \hat{o}^s)$. (iii) Action-reason retrieval: LLM uses the combination of (i) and (ii) to retrieve final action-reason statement. 
consisting of three steps: (i) Atypicality-aware image verbalization: Using LLaVA \cite{liu2023visual} and LLM, we generate a coherent verbalization $\mathcal{T}_{\mathcal{V}}$ sensitive to atypicality; (ii) Atypicality statement detection: We detect the atypicality statement $\hat{s}=(\hat{a}, \hat{o}^p, \hat{o}^s)$; (iii) Action-reason retrieval: We retrieve the final action-reason statement using (i) and (ii). Fig.~\ref{fig:method-overview} illustrates the proposed method. 

% \noindent

%\textbf{(i) Atypicality-aware image verbalization.}
\textbf{(i) \rev{Verbalize image in atypicality-aware manner.}} 
Each ad image $I=\left(V, T\right)$ is composed of visual content ($V$) (objects) and textual content ($T$) (scene-text). We obtain $V$ and $T$ by querying LLaVA for up to 5 objects and a list of text-scenes visible in the image. However, this information is insufficient to fully comprehend the image. Hence, we additionally prompt LLaVA with
(1) $ImageNarration$ ($IN$): LLaVA's responses when prompted \textit{Describe the image in detail.} (2) $UnusualHighlighter$ ($UH$): LLaVA's reponse when asked \textit{What is unusual about the image?}.
MAC results (Table~\ref{tab:MAC_ASR_Full_set}) show that $UH$ effectively captures image unusualness (closely related to atypicality), while $IN$ provides scene and object information useful for retrieving atypicality statement $\hat{s}$. Thus, we construct the final verbalization $\mathcal{T}_{\mathcal{V}}$ by combining $UH$, $V$, $T$, and $IN$ using an LLM.

% Thus, to construct the final description $\mathcal{T}_{\mathcal{V}}$, we propose combine unusualness (i.e., $UH$) with objects $V$, scene-text $T$, and image description ($IN$). We leverage an LLM to combine all this information.
%in a monolithic manner. 
%$\mathcal{T}_{\mathcal{V}}$ will be further exploited to detect the action-reason statements. 

% \noindent
%\textbf{(ii) Atypicality Statement Retrieval (ASR).}
\textbf{(ii) \rev{Detect atypicality statement.}}
We construct all possible statements to predict the atypicality statement $\hat{s}$. Option Generator (\textcolor{teal}{teal} module in Fig~\ref{fig:method_asd}) combines $V$ and $\mathcal{S}_{\mathcal{A}}$ to generate all possible statements ${S}_I$. Specifically, each object pair $(o_i,o_j) \in V$ is combined with all atypicality statement template $s \in \mathcal{S}_{\mathcal{A}}$ to create atypicality statements 
%$(a_k, o_i/o_j, o_j/o_i)$ 
\rev{$(a_k, o_i, o_j)$ and $(a_k, o_j, o_i)$}
for all $a_k \in \mathcal{A}$. We then pass these atypical statements ${S}_I$ along with verbalization $\hat{S}_{IN}$%$\mathcal{T}_{\mathcal{V}}$ 
into the classifier to predict $\hat{s}$ (no ground-truth is used). 

% \noindent
%\textbf{(iii) Action-Reason Retrieval (ARR)}. 
\textbf{(iii) \rev{Retrieve action-reason statements.}}
\cite{hussain2017automatic} provided three action-reason statements for each image, each offering different plausible reasons for the same action. Given these three plausible (i.e., positive) and many implausible (i.e., negative) action-reason statements, the task is to detect all plausible action-reason statements $\{\hat{ar}_I \}_{i=1}^3$. 
We proposed various verbalization strategies, including concatenation and LLM-based combinations (i.e., $\mathcal{T}_{\mathcal{V}}$) of $V, T, IN, UH$, as well as concatenation of $\hat{s}$ with $\mathcal{T}_{\mathcal{V}}$, to be utilized by an LLM for retrieving the final action-reason statement.
% We proposed different verbalization variants, including concatenation and LLM-based combination (i.e. $\mathcal{T}_{\mathcal{V}}$) of $V, T, IN, UH$, as well as concatenation of $\hat{s}$ with $\mathcal{T}_{\mathcal{V}}$, to be leveraged by an LLM %GPT-4 
% to retrieve the final action-reason statement. %Furthermore, to demonstrate that our approach is capable of a deep understanding of the image, we leveraged an LLM to produce \textbf{semantically hard negatives} (Sec.~\ref{sec:exp_setup}).

% \input{sections/ECCV_dumps/method}
\section{Experimental Setup}
\label{sec:exp_setup}

\textbf{Datasets.}
PittAds~\cite{hussain2017automatic} includes 64,832 ad images, with 3,928 annotated for atypicality, action-reason, and primary/secondary objects. For MAC, we use atypicality categories, while for AOR and ASR, we use primary/secondary objects along with our atypicality statement templates $\mathcal{S}_\mathcal{A}$ (Fig.~\ref{fig:atyp_def_stmt}) to generate ground truth for evaluation. We utilized the train set with
% from the challenge
% \footnote{https://evalai.cloudcv.org/web/challenges/challenge-page/86/overview)} 
% with 
1,168 samples for the main results, as it includes at least one annotation of the atypicality categories studied and is larger than the test set. No training was performed. Ablation studies are reported on a smaller subset of 250 images due to the high cost.

\textbf{Baselines.} 
We use LLaVA (`vicuna-13b-v1.5') \cite{liu2023visual}, InstructBLIP (`vicuna-13b-v0')~\cite{dai2023instructblip}, MiniGPT4 \cite{zhu2024minigpt} (`vicuna-13b-v0'), and CLIP~\cite{radford2021learning} (`ViT-L/14@336px' following \cite{jia-etal-2023-kafa}), and `InternVL-Chat-V1-1' (denoted as InternVL-V1) \cite{Chen_2024_CVPR} as VLM baselines (and `InternVL2-8B', LLaVA 1.6 in supp Tab.~4). LLaVA is our multimodal component due to its GPT-4-informed instruction tuning, state-of-the-art reasoning, and promptability \cite{liu2023visual}. We evaluate GPT-4V on a limited 250 examples, constrained by cost. We report BLIP-2~\cite{li2023blip} (`blip2-flan-t5-xl') only for AOR as it failed to produce meaningful output for multi-option tasks (i.e., MAC and multi-ARR; detail in supp sec.~2.2).
% when asked to return multiple options, i.e., MAC and multi-ARR (detail in supp sec.~2.2).

Our analysis spans recent public LLMs, such as `vicuna-13b-v1.5' (Vicuna)~\cite{Touvron2023Llama2O}, `InternLM2-5‑7b‑chat' (InternLM; see supp sec.~2.2), and leading commercial models like GPT-3.5/4. We chose Vicuna as it is used in all VLMs (LLaVA, MiniGPT4, InstructBLIP) and InternLM as InternVL2-8B's LM. We also compare GPT-4 and GPT-4V. We introduce CLIP ($I+T$) as a zero-shot baseline aligned with KAFA \cite{jia-etal-2023-kafa} but avoid direct comparison with KAFA as it is not publicly available. To assess our atypicality method, we compare against $V+T$ (verb. baseline), which includes basic image information (up to 5 objects, scene text). 

\textbf{Metrics.} 
We use Precision, Recall, and F1-score 
to evaluate MAC %to address data imbalance
(additional metrics in supp Tab.~1). For AOR, we assess sentence similarity between $s^{+}$ and $\hat{s}$ using `all-mpnet-base-v2' \cite{song2020mpnet}, a state-of-the-art sentence embedding method. Common text-matching and accuracy metrics aren't suitable for AOR since it is a generative task, and annotations can vary widely (e.g., `beer,' `glass of beer,' `beer glass') due to human inconsistencies and typos. We report accuracy (denoted `Acc') for single statement retrieval tasks, where the model returns only one statement per query (i.e., ASR and Single ARR). Top-k Acc and unranked Precision@k are the metrics for the multi-option ARR, with $Precision@k = \frac{min(k,\textit{\# of relevant statements in top k predictions})}{k}$. Note top-3 acc and prec@1 are the same. 

\textbf{Hard Negative (HN) Generation.} 
To measure ARR performance, \cite{ye2018advise} mined hard negatives from images within the same topic%(e.g., `clothing')
, while \cite{jia-etal-2023-kafa} expanded the negative options. These negatives often include irrelevant objects, allowing VLMs to easily disregard them by comparing objects. This hinders accurate measurement of models' reasoning ability.
% different from those in ground-truth action-reason statements. This allows VLMs to easily disregard them by comparing objects, hindering accurate measurement of the model's reasoning ability.
In contrast, a concurrent work \cite{bavaresco-etal-2024-dont} used annotators to write implausible statements based on visible objects/texts. Our approach differs in three key ways: (1) it is LLM-based, image-agnostic, and scalable; (2) it generates a wider variety of negatives while focusing on semantics (e.g., altering actions, reasons, adjectives, or swapping objects not visible such as `lipstick' instead of `lip balm'), and (3) they evaluate contrastive-based VLMs, whereas we focus on generative VLM/LLMs with stronger reasoning ability.

Specifically, for each ground-truth action-reason statement, we ask GPT-4 to generate a negative action-reason statement by (1) \textbf{Action alter}: changing the action to an unrelated or opposite action while preserving the reason; (2) \textbf{Reason alter}: changing the reason to an unrelated or opposite reason; (3) \textbf{Adjective alter}: negating or modifying adjectives to make the statement incorrect; (4) \textbf{Object swap}: substituting at least one object in the statement with an unrelated object; (5) \textbf{Statement alter}: generating a completely unrelated action-reason statement.

We validated our hard negatives by sampling 100 images and asked human annotators to select all correct statements (results in supp sec.~2.2).

% To address this limitation, we introduce a novel method of creating semantically challenging negatives through LLMs. For each ground-truth action-reason statement, we ask GPT-4 to generate a negative action-reason statement by: (1) \textbf{Action alter}: altering the action to an unrelated/opposite action, while preserving the reason, (2) \textbf{Reason alter}: altering the reason to an unrelated/opposite reason, (3) \textbf{Adjective alter}: negating or altering adjectives to make the statement unrelated or incorrect, (4) \textbf{Object swap}: modifying at least one object in the statement to one not used in the statement, (5) \textbf{Statement alter}: generating an action-reason statement that is completely unrelated to the ground-truth. 

% Table~\ref{tab:ablation_hard_easy_negative} reveals that while VLM performance significantly decreases with these semantically hard negatives, our method maintains robust performance without additional training.

\textbf{Implementation.}
For GPT-4, GPT-3.5, and Vicuna temperature is set to 0. For LLaVA \cite{liu2023improvedllava}, BLIP-2 \cite{li2023blip}, \rev{InternVL models \cite{Chen_2024_CVPR} and InternLM \cite{cai2024internlm2} we used the original setting}. We applied 8-bit quantization for LLaVA, MiniGPT-4, Vicuna, InternVL and InternLM. All experiments were zero-shot and conducted on an NVIDIA Quadro RTX A5000. Example prompts are in supp.

% \textbf{Experimental Details.}

% \textit{MAC.}

% \textit{ASR.}

% \textit{AOR.}
% We provide the true atypicality relation $a^+$ without the primary/secondary objects to the model $s = (a^{+}, -, -)$. To aid with the output format, the model is further provided an example $s^{eg} = (a^{+}, o_{p}^{eg}, s_{s}^{eg})$ sampled from the same atypicality relation. The model generates $s^{pred} = (a^{+}, o_{p}^{pred}, s_{s}^{pred})$ with the predicted primary/secondary objects. To evaluate this against the ground truth $s^{+} = (a^{+}, o_p^{+}, o_s^{+})$, we obtain a sentence similarity score between $s^{+}$ and $s^{pred}$.

% \textit{ARR}

% We provide the true atypicality without the primary/secondary objects to the model, i.e., $s = (a^{+}, -, -)$, and obtain $s^{pred} = (a^{+}, o_{p}^{pred}, s_{s}^{pred})$ with the predicted primary/secondary objects. To evaluate this against the ground truth $s^{+} = (a^{+}, o_p^{+}, o_s^{+})$, we obtain a sentence similarity score between $s^{+}$ and $s^{pred}$. For our results in \ref{tab:atyp-obj-recog}, we use all-mpnet-base-v2 \cite{song2020mpnet} for computing sentence similarity.

% \input{prompts/aor_prompt}

\section{Results}
\label{sec:results}

The key goal is to benchmark VLMs and evaluate their understanding of persuasive ads. This section first presents results on VLMs/LLMs' atypicality understanding tasks,
%(Sec~\ref{sub-sec:results_atypic_understanding}), 
forming the foundation for our atypicality-aware verbalization. 
Then, we explore whether atypicality can help ad understanding on Action-Reason Retrieval (ARR).
%Then, we evaluate VLMs/LLMs' zero-shot reasoning ability on the ARR task and explore whether atypicality can help advertisement image understanding on the Action-Reason Retrieval (ARR) task. 
% n Sec~\ref{subsec:action_reason_results} and Sec~\ref{subsec:analysis_ablation}. 
% All results are on full-set otherwise stated.

\subsection{Atypicality Understanding Results}
\label{sub-sec:results_atypic_understanding}
We assess open-source VLMs' (LLaVA~\cite{liu2023improvedllava}, InstructBLIP~\cite{dai2023instructblip}) understanding of the atypicality and GPT-4V. Additionally, we evaluate the effectiveness of two prompting strategies for capturing atypicality: (1) $IN$ (\textit{Describe the image in detail.}) and (2) $UH$ (\textit{What is unusual about the image?}). These strategies are compared to the VLMs and the $V+T$ baseline across three LLMs (GPT-4, GPT-3.5, Vicuna). Table~\ref{tab:MAC_ASR_Full_set} summarizes the results on MAC and ASR, while AOR results are in Table~\ref{tab:atyp-obj-recog-full}.
% For fair assessment of VLMs zero-shot generalizability, all models are in zero-shot. 

\setlength{\tabcolsep}{2pt}
\begin{table}[t] %[!tp]
\centering

\scriptsize
\begin{tabular}{lc||cccccc||c}
\toprule
& &\multicolumn{6}{c||}{\textbf{\textit{MAC}}} &\textbf{\textit{ASR}} \\
% \cmidrule(lr){3-8}
% \cmidrule{9}
\textbf{Classifier} & \textbf{Method} &\multicolumn{2}{c}{\textbf{Precision}} &\multicolumn{2}{c}{\textbf{Recall}} &\multicolumn{2}{c||}{\textbf{F1-score}} & \textbf{Acc} \\
% \cmidrule{3-8}
& &$\checkmark$ &$\times$ &$\checkmark$ &$\times$&$\checkmark$ &$\times$ & \\
\hline
\hline
\multirow{3}{*}{LLaVA \cite{liu2023visual}}  &I &27.75 &27.75 &42.38 &52.71 &21.24 &26.03 &  18.83 \\
&IN &25.12 &31.40 &42.44 &53.04 &25.06 &31.32& 20.90 \\
&UH &44.35 &30.44 &42.04 &52.44 &24.16 &29.98 & 17.90 \\
InstructBLIP \cite{li2023blip}&I &34.81 &27.60 &41.43 &50.73 &17.72 &20.18 & 19.76 \\
% MiniGPT-4 \cite{zhu2024minigpt}&- &0.00 &0.00 &0.00 &0.00 &0.00 &0.00 &0.00 \\
\hline
\hline
\multirow{3}{*}{Vicuna~\cite{vicuna2023}} 
&$V + T $ &36.70&30.64  &41.73 &45.78 &32.52 &31.66 & 14.30 \\
&$IN$& 37.71 & 32.04 & \textbf{43.70} & \textbf{45.91} &\textbf{ 34.51} &\textbf{ 32.09} & \textbf{23.29}\\
&$UH$&\textbf{39.41} &\textbf{33.33}  &36.05 &42.88 &27.35 &30.36 &14.74\\
\hline
\multirow{3}{*}{GPT 3.5} &$V + T$&41.46 &35.36 &23.21 &21.54 &28.18 &24.95 &50.00 \\
&$IN$& 46.28 & 42.50 & 25.13 & 14.75 & \textbf{28.49} & 19.64 & \textbf{50.55} \\
&$UH$ &\textbf{49.10} &\textbf{43.34} &\textbf{27.38} &\textbf{30.92} &27.06 &\textbf{28.24} &50.05 \\
\hline
\multirow{2}{*}{GPT 4} &$V + T$ &40.38 &35.95 &22.56 &6.69 &22.66 &10.99 &52.44 \\
&$IN$& \textbf{54.78} & \textbf{53.40} & 27.19 & 13.64 & 30.58 & 20.91 & \textbf{57.70} \\
&$UH $&53.49 &51.01 &\textbf{29.15} &\textbf{28.89} &\textbf{34.62} &\textbf{33.05} &56.89 \\
\bottomrule
\end{tabular}
\caption{\textbf{Atypicality Understanding Tasks (MAC \& ASR) on Full-set}. $UH$ is very effective on MAC.
%for extracting unsualness that is very close to the atypicality. 
$IN$ slightly outperforms $UH$ on ASR; ASR also requires to identify the objects which may not be well-represented in $UH$. $\checkmark$/$\times$ denotes performance with/without No Atypicality (NA) class. Best result per LLM bolded.}
\label{tab:MAC_ASR_Full_set}
\vspace{-5pt}
\end{table}

    \textbf{VLMs struggle with direct atypicality inference}.
    % Table~\ref{tab:MAC_ASR_Full_set}compares atypicality information extraction strategies ($UH$ and $IN$) against VLM and $V+T$ baselines on MAC and ASR tasks.
    Table~\ref{tab:MAC_ASR_Full_set} shows VLMs consistently underperform verbalization approaches ($UH$ and $IN$) in both MAC (F1-score) and ASR. For instance, LLaVA shows high recall (52.71) but low precision (27.75) in the MAC task, indicating it over-predicts atypicalities. This trend is particularly evident in the OIO and TR2 categories, where LLaVA achieves perfect and near-perfect recall (1.0 and 0.79) but low precision (0.18 and 0.23). Conversely, recall for other categories doesn't exceed 0.24, suggesting a bias towards OIO and TR2 predictions (category-wise metrics not shown in table).
    InstructBLIP exhibits similar issues. 
    This suggests VLMs lack the reasoning ability to accurately infer atypicality.
    %, instead broadly predicting atypicalities without discernment. 
    
    % This can be confrimed by LLaVA's high  recall on `OIO' and `TR2' (1.0 and 0.79, respectively), while achieving low precision (0.18 and 0.23,respectively). LLaVA's recall on other atypical categories does not exceed 0.24, suggesting over prediction of `OIO' and `TR2' for most images.  

\textbf{\textit{V+T} lacks sufficient context to fully understand the image.}
$V+T$ provides inadequate context for extracting atypicality, as evidenced by GPT-4's low recall/F1 (precision/recall scores of 58.11/86.04 for NA) and low recall for atypical categories (6.00, 0.79, 6.10, and 13.86 for TR1, TR2, OIO, and OR; not shown in table). Smaller LLMs (Vicuna and GPT-3.5) perform better than GPT-4 with V+T, but their improvement is likely due to hallucination rather than extracting useful information. In contrast, $UH$ and $IN$ provide richer descriptions that better capture atypicality.

\textbf{VLMs are effective for verbalization}.
In MAC, $UH$ emerges as the top-performing strategy, effectively extracting unusualness and atypicality from images. It significantly improves $V+T$ F1 score without NA (details in section~\ref{sec:atyp-tasks}) by 3.29 and 22.06 for GPT-3.5 and GPT-4, respectively, and is only slightly lower than Vicuna's $V+T$. $UH$ also outperforms $IN$ overall. In ASR, $UH$ is more effective than $V+T$ and outperforms VLMs by a significant margin when used in GPT models (e.g., by 37.13 with GPT-4 compared to the strongest VLM). However, $IN$ slightly outperforms $UH$ on GPT-4/GPT-3.5 and significantly on Vicuna. These results indicate that identifying atypicality requires both image understanding and reasoning capabilities. ASR necessitates the model to identify the correct atypicality statement, which includes both objects and their relation (e.g., \underline{search bar} completely replaces \underline{mouth} in its usual context, assuming its function or position in Fig.~\ref{fig:intro_examples} (a)).
Thus, $IN$ may be a better verbalization as it provides more detailed information about the image, objects, and their relations, whereas $UH$ offers implicit information about the image's unusualness. 
Therefore, we use $UH$ (best on MAC) and $IN$ (best on ASR) along with $V+T$ to create atypicality-aware verbalization. Inspired by ASR results we also adopt $IN$ for detecting atypicality statement $\hat{s}_{IN}$ which is combined with our verbalization for ARR.% instead of just categories.

%\begin{adjustwidth}{-4.5 cm}{-4.5 cm}\centering
\begin{table}[t]%[!htp]
\centering
\scriptsize

\begin{tabular}{l<{\hspace{0.25em}}||>{\hspace{0.25em}}c|>{\hspace{0.5em}}c>{\hspace{1em}}c>{\hspace{1em}}c>{\hspace{1em}}c}

\toprule
\multirow{2}{*}{\textbf{Model}} & {\textbf{\textit{Avg. }similarity} ($\hat{s}$ to $s^+$)} & \multicolumn{3}{c}{\textbf{\textit{\% of scores}}} \\
& \textbf{score} & \textbf{$>$ 0.7} & \textbf{$>$ 0.6}  & \textbf{$>$ 0.5 } \\
\hline
\hline
BLIP2 \cite{li2023blip} & 0.45 & 8.77 & 19.78 & 35.43 \\
InstructBLIP \cite{dai2023instructblip} & 0.46 & 9.54 & 21.24 & 40.76 \\
MiniGPT4 \cite{zhu2024minigpt} & 0.51 & 15.24 & 32.28 & 51.71 \\
LLaVA \cite{liu2023visual} & \textbf{0.59} & \textbf{31.41} & \textbf{51.35} & \textbf{65.16} \\
\hline
GPT-4V $\dagger$ \cite{2023GPT4VisionSC} & 0.67 & 46.94 & 61.63 & 77.14 \\
\bottomrule
\end{tabular}
\caption{\textbf{AOR on Full-set}. 
%MPNet \cite{song2020mpnet} sentence similarity scores between $s^{pred}$ and $s^+$, and score thresholds. \\ 
$\dagger$ on Small-set.}
%See supplement for the complete results on small-set.}
\label{tab:atyp-obj-recog-full}
% \vspace{-5pt}
\end{table}
%\end{adjustwidth}

% \textbf{Atypical Object Recognition.}
\textbf{VLMs' limitations in Atypical Object Recognition}.
Table~\ref{tab:atyp-obj-recog-full} explores VLMs performance on the AOR task, which involves generating primary and secondary objects given only GT atypicality relation to complete its atypicality statement from Fig.~\ref{fig:atyp_def_stmt}.
% Sentence similarity scores between $\hat{s}$ and $s^+$ were computed using MPNet \cite{song2020mpnet}.
Scores above 0.7 indicate strong semantic overlap, while scores between 0.5 and 0.7 indicate moderate overlap.
The results underscore current VLMs' difficulty in reasoning and recognizing atypical objects. InstructBLIP and MiniGPT4 perform poorly, with most predictions scoring below 0.5, highlighting their struggle to recognize primary and secondary objects in atypical contexts. GPT-4V emerges as the most proficient model, yet only about half of its predictions surpass the 0.7 mark. These results highlight the need to improve VLMs' reasoning in complex visual scenes, such as atypical objects. 

% for further advancements in VLM development to enhance their reasoning skills in complex visual scenarios such as atypical objects.

\subsection{Action-Reason Retrieval Results}
\label{subsec:action_reason_results}

\begin{figure}[!tp]
% \noindent % Ensures we start from the very left of the page
\hspace{-10pt}
\begin{minipage}{0.22\textwidth}
\vspace{-10pt}
\centering
% \vspace{-10pt}
\includegraphics[width=\linewidth]{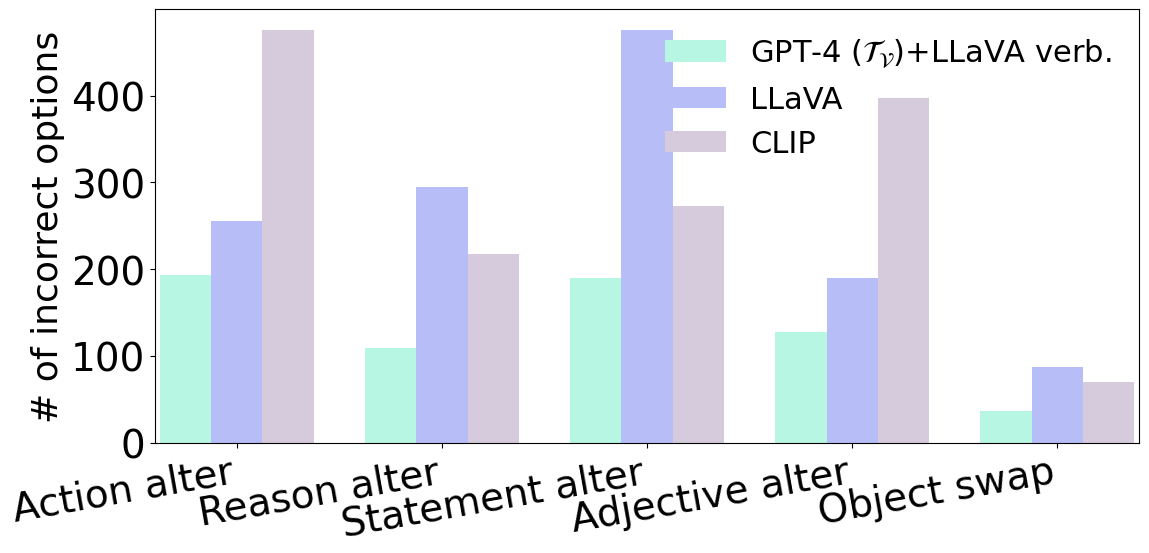}
\captionof{figure}{\textbf{ARR error analysis on Full-set}. }%GPT4($\mathcal{T}_\mathcal{V}$) makes significantly less semantic errors than VLMs.}
\label{fig:ARR_Error_Analysis}
\end{minipage}
% \hfill % Fill the space between the figure and table
\hspace{-2pt}
\begin{minipage}{0.26\textwidth}
\vspace{-10pt}
\centering
\scriptsize
\begin{tabular}{l||ccc|cc}\toprule
% \multirow{3}{*}{\textbf{Classifier}}&\multicolumn{6}{c}{\textbf{\textit{Multi}}}  \\ %& \textbf{\textit{Single}}\\

\multirow{2}{*}{\textbf{Classifier}}&  \multicolumn{3}{c}{\textbf{\textit{precision@k}}} & \multicolumn{2}{c}{\textbf{\textit{Top-k acc}}}\\     %&\multirow{2}{*}{\textbf{Acc}} \\
 & \textbf{k=1} &\textbf{k=2 }&\textbf{k=3} &\textbf{k=1} &\textbf{k=2} \\
\hline
\hline
LLaVA  &59.67 &38.27 &26.06 &32.92 &48.14 \\ %&26.00 \\
\hline
GPT-4V  &97.17 &89.91 & 74.86 & 77.01 & 90.32 \\ %&90.00\\
% GPT-4($\mathcal{T}_\mathcal{V}$ )&\textbf{97.58} &\textbf{90.72} &\textbf{76.61} &\textbf{81.04} &\textbf{92.74} &\textbf{97.58} \\ %&97.90\\
% GPT-4&\textbf{97.58} &\textbf{90.72} &\textbf{76.61} &\textbf{81.04} &\textbf{92.74} &\textbf{97.58} \\ %&97.90\\
% GPT-4  $\mathcal{T}_\mathcal{V} + \hat{s}_{\mathcal{T}_\mathcal{V}}$&\textbf{97.97} &90.49 &75.70 &\textbf{81.37} &94.73 &97.97 &0 \\
GPT-4 ($\mathcal{T}_\mathcal{V}$)&\textbf{97.58} &\textbf{90.72} &\textbf{76.61} &\textbf{81.04} &\textbf{92.74} \\ %&97.90\\
\bottomrule
\end{tabular}
\captionof{table}{\textbf{GPT-4V verb. 
% for $\mathcal{T}_{\mathcal{V}}$ 
in Multi-ARR on Small-set}. }
\label{tab:action_reason_gpt4v}
\end{minipage}
% \vspace{-3pt}
\end{figure}

We evaluate our proposed method using three LLMs (GPT-4, GPT-3.5, Vicuna, \rev{InternLM in supp}) and compare them against VLMs (CLIP, LLaVA, InstructBLIP, InternVL-V1, GPT-4V, and LLaVA1.6 and InternVL2-8B in supp.). %Additionally, we benchmark against the closest zero-shot version of KAFA~\cite{jia-etal-2023-kafa}. 
Table~\ref{tab:full_multi_selection_action_reason} presents the multi-option action-reason retrieval results (ARR) and their comparison against VLMs on the full-set. Table~\ref{tab:action_reason_gpt4v} demonstrates ARR results on small-set when GPT-4V is the VLM in our pipeline. We adopted Vicuna as the public LLM as it is used in all VLMs (LLaVA, InstructBLIP, and MiniGPT4). We use GPT-4 and GPT-4V as powerful LLM/VLM pairs.

\begin{table}[t]%[!htp]
\centering
\scriptsize
\setlength{\tabcolsep}{1pt}
\begin{tabular}{lc||ccc||cc|l}
% \hline
\toprule
\multirow{2}{*}{\textbf{Classifier}} &\multirow{2}{*}{\textbf{Verb.}} & \multicolumn{3}{c||}{\textit{\textbf{Precision@k}}} & \multicolumn{2}{c|}{\textit{\textbf{Top-k Acc}}} & \cellcolor{LGray}  \textbf{Avg}\\

& &\textbf{k=1}&\textbf{k=2} &\textbf{k=3} &\textbf{k=1}&\textbf{k=2}&\cellcolor{LGray}\\
\hline
\hline
 \multirow{2}{*}{CLIP\cite{radford2021learning}}   &$I$ & \textbf{61.04} & 33.86 & 22.66 & 23.72 & 44.61 & \cellcolor{LGray} 37.18\\
  & $I+T$&46.15 &24.36 &16.24 & 15.15 & 31.25 & \cellcolor{LGray}  26.63\\
  % & $I+V+T$ &\textbf{70.46} & \textbf{39.17} & \textbf{26.11} & \textbf{29.79} &\textbf{ 53.08} & \cellcolor{LGray} \textbf{43.72}\\ # TODO move this to supplement
 
\multirow{2}{*}{LLaVA~\cite{liu2023visual} } &$I$ &59.67 &\textbf{38.27} & \textbf{26.06} & \textbf{32.92} &\textbf{ 48.14} & \cellcolor{LGray}  \textbf{41.01}\\
&$I + \mathcal{T}_\mathcal{V}$ &59.45 &37.37 &25.14 & 27.49 & 47.07 & \cellcolor{LGray} 39.30\\
InstructBLIP~\cite{dai2023instructblip} &$I$ &15.05 & 10.03 & 7.80 & 13.04 & 13.04 & \cellcolor{LGray} 11.79 \\
InternVL-V1~\cite{Chen_2024_CVPR} & $I$ & 52.22 & 32.79 & 22.17 & 22.51 & 40.66 & \cellcolor{LGray} 30.07 \\
 
% MiniGPT4~\cite{zhu2024minigpt} &$I$& 0.00 & 0.00 & 0.00 & 00.00 & 00.00 & 00.00\\
 \hline
 \hline
\multirow{5}{*}{Vicuna~\cite{vicuna2023}}&$V + T$& 64.13 & 40.71 &27.57 & 21.49 & \textbf{43.41} & \cellcolor{LGray} 39.46\\
 
 % &$\mathcal{T}_\mathcal{V}+\hat{s}_{UH}$ with GPT-4 Verb.& \textbf{69.09} & \textbf{45.16} & \textbf{30.71} & \textbf{24.49} & \textbf{43.84} & \textbf{69.09 } \\
     &$\mathcal{T}_\mathcal{V}$ (Ours) & 67.38 & 44.01 & 29.94  & 23.20 & 41.95 & \cellcolor{LGray} 41.30 \\
   &$\mathcal{T}_\mathcal{V} + \hat{s}_{IN}$ (Ours) & 68.32 & \textbf{44.52} &30.25&22.95& 43.24 & \cellcolor{LGray} 41.86\\
    &$\mathcal{T}_\mathcal{V}$ (GPT-4 Verb.) (Ours) & \textbf{68.49} & \textbf{44.52} & \textbf{30.37}  & \textbf{24.06} & 43.24 & \cellcolor{LGray} \textbf{42.14} 
 % #TODOD
 \\\hline
% \multirow{3}{*}{GPT 3.5} &$V + T $& 0.00 & 0.00 & 0.00 & 00.00 & 00.00 & 00.00\\
%  &$\mathcal{T}_\mathcal{V}$ & 0.00 & 0.00 & 0.00 & 00.00 & 00.00 & 00.00\\
%  &$\mathcal{T}_\mathcal{V} + \hat{s}_{UH}$ & 0.00 & 0.00 & 0.00 & 00.00 & 00.00 & 00.00\\
 % \hline
\multirow{4}{*}{GPT-4} &$V+T$&93.73 &84.42 &70.50  & 71.50 & 89.87 & \cellcolor{LGray} 82.00\\
 &$\mathcal{T}_\mathcal{V}$ (Ours) &93.99 &86.35 &72.96 & 74.94 & 91.16 & \cellcolor{LGray} 83.88 \\
 % &$\mathcal{T}_\mathcal{V} + \hat{s}_{UH}$&94.42 &85.95 &72.31  & 74.31 & 91.84 & 94.42\\
 & $\mathcal{T}_\mathcal{V} + \hat{s}_{IN}$ (Ours) & \textbf{95.54} &\textbf{87.55} &\textbf{74.62}  & \textbf{88.42}& \textbf{93.40} & \cellcolor{LGray} \textbf{87.91}\\
\bottomrule
\end{tabular}
\caption{\textbf{ARR on Full-set
}. Predicted atypicality statement $\hat{s}$ uses the respective prompting (IN) and LLM (Vicuna, GPT-4), with all LLMs using LLaVA verbalization. $\mathcal{T}_\mathcal{V}$ combines $V+T+IN+UH$ using Vicuna or GPT-4. Best number per block and column is bolded. Task is Multi-ARR. Precision@k is unranked.
}
\label{tab:full_multi_selection_action_reason}
\end{table}

\textbf{LLMs are more powerful than VLMs.}
In Table~\ref{tab:full_multi_selection_action_reason} and \ref{tab:action_reason_gpt4v}, LLMs with atypicality-aware verbalization ($\mathcal{T}_\mathcal{V}$) consistently outperform VLMs. For example, GPT-4 surpasses LLaVA by 42.87 points. GPT-4 also outperforms a strong VLM, GPT-4V, in all metrics (Table~\ref{tab:action_reason_gpt4v}). \rev{Results in supp confirm this trend across InternVL2-8B and LLaVA~1.6.} \textit{This highlights the superior reasoning ability of LLMs in understanding atypicality and action-reason statements.} 

Interestingly, when LLaVA is provided with both image and $\mathcal{T}_\mathcal{V}$ (LLaVA($I +\mathcal{T}_\mathcal{V}$)), it underperforms Vicuna by 7.93, 6.64, and 4.8 points on prec@1, prec@2, and prec@3, respectively. Also, its performance drops by 1.71 points compared to LLaVA($I$). \textit{This reveals that despite using Vicuna as its LLM, LLaVA's reasoning ability is limited, hindering it effective use of $\mathcal{T}_\mathcal{V}$ for action-reason retrieval}. 

% In tasks requiring multi-step reasoning like ARR, LLaVA's performance drops when using $\mathcal{T}_\mathcal{V}$. We conjecture this is due to VLMs' lower reasoning ability than LLMs, even those using similar models (i.e., Vicuna is higher than LLaVA($I+\mathcal{T}_{\mathcal{V}}$) by 4.48 on prec@3\
% ).
% This may be attributed to the simplicity of VL pre-training data and tasks.

% While advantageous for action-reason detection, CLIP's approach limits its applicability in tasks like action-reason generation and reasoning.

\setlength{\tabcolsep}{0.5pt}

\textbf{Atypicality boosts persuasive visual media understanding.} 
% Table~\ref{tab:full_multi_selection_action_reason} demonstrates the effectiveness of our atypicality-aware verbalization method compared to VLMs and the basic $V+T$ verbalization baseline.
Table~\ref{tab:full_multi_selection_action_reason} highlights the effectiveness of our atypicality-aware verbalization compared to VLMs and the basic $V+T$ baseline. 
Vicuna($V+T$) falls 1.55 points behind LLaVA on \textit{avg}, \textit{reflecting insufficient context from basic verbalization}.
In contrast, $\mathcal{T}_\mathcal{V}$ consistently improves $V+T$, with gains of 3.8/2.8 (prec@2/prec@3) for Vicuna and 1.93/2.46 for GPT-4. 
Combining $\mathcal{T}_\mathcal{V}$ with $\hat{s}_{IN}$ achieves the best results across all LLMs (GPT3.5 in supp), confirming the benefit of incorporating atypicality-aware verbalization and atypicality.
% While Vicuna($V+T$) lags behind LLaVA by 1.55 points on \textit{avg}, \textit{indicating insufficient context from basic verbalization}, our $\mathcal{T}_\mathcal{V}$ consistently improves $V+T$ performance. We observe gains of 3.8/2.8 and 1.93/2.46 on prec@2/prec@3 for Vicuna and GPT-4, respectively.  
% Combining $\mathcal{T}_\mathcal{V}$ with $\hat{s}_{IN}$ achieves the best results across all LLMs (GPT3.5 in supp), confirming the benefit of incorporating atypicality-aware verbalization and atypicality.
Notably, Vicuna shows minimal improvement with GPT-4's verbalization, indicating our strategy does not strictly rely on extensive LLMs like GPT-4.

\begin{figure}[!tp]
% \noindent % Ensures we start from the very left of the page

\begin{minipage}{0.22\textwidth}
\vspace{-18pt}
% \hspace{-10pt}
% \centering
\includegraphics[width=\textwidth]{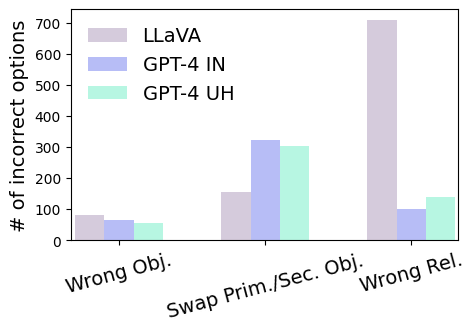}
    \caption{\textbf{ASR error analysis on Full-set}}
    \label{fig:ASR_Error_Analysis}
\end{minipage}
% \hfill % Fill the space between the figure and table
% \hspace{-0pt}
% \hspace{-17pt}
\begin{minipage}{0.25\textwidth}
\vspace{-5pt}

\centering
\scriptsize
\begin{tabular}{l||cccc}\toprule
 \textbf{Model }& \textbf{$I$}& \textbf{$I+s^{+}$}& \textbf{$I+\hat{s}$}& \textbf{$I+\Bar{s}$} \\
 \hline
 \hline
 LLaVA & 26.00 & 35.18 & \textbf{54.28} & 28.16 \\
 InstructBLIP & 20.44 & 23.25 & \textbf{23.40} & 19.69\\
 GPT-4V & 86.87 & \textbf{89.35} & 87.24 & 86.96 \\
 \hline
 GPT-4 ($\mathcal{T}_{\mathcal{V}}$)& \textbf{96.77} & 91.42 & 96.76 & 90.20 \\
\bottomrule
\end{tabular}
\captionof{table}{\textbf{Atypicality ablation on ARR Small-set.} 
$I$: image, $s^+$: GT atyp., $\hat{s}$: predicted atyp. based on $\mathcal{T}_\mathcal{V}$, $\Bar{s}$: incorrect atyp. Acc on Single ARR. Best result per row in bold.}
\label{tab:atypicality_ablation_vlm}
\end{minipage}
\end{figure}

Table~\ref{tab:atypicality_ablation_vlm} investigates how the addition of atypicality statements impacts VLM performance on ARR using three methods: (1) $I+s^+$, with true atypicality statements  $s^+ = (a^+, o^{+p}, o^{+s})$; (2) $I+\hat{s}$, with predicted atypicality statements $\hat{s} = (\hat{a}, \hat{o}^p, \hat{o}^s)$ by GPT-4 and $\mathcal{T}_\mathcal{V}$; and (3) $I+\Bar{s}$, with incorrect atypicality statements using correct objects but incorrect relations, $\Bar{s} = (\Bar{a}, o^{+p}, o^{+s})$. The results reveal that VLMs benefit from atypicality statements. However, 
for GPT-4 (an LLM), adding the atypicality to $\mathcal{T}_{\mathcal{V}}$ is not as useful, since $\mathcal{T}_{\mathcal{V}}$ already contains correct atypicality information, and incorrect statement reduce the performance. These findings highlight the importance of incorporating atypicality to improve VLM performance on ARR tasks.

\textbf{Generalization to typical images.} We compared our pipeline with Vicuna ($\mathcal{T}_{\mathcal{V}}$) on typical images (i.e., no atypicality) against LLaVA on small-set ARR (details in supp). Vicuna ($\mathcal{T}_{\mathcal{V}}$)
achieved 71.2\%/48.6\%/33.2\% vs. 66.4\%/42.2\%/28.3\% for LLaVA on prec@1/2/3, demonstrating our approach's effectiveness beyond atypical ads. 

% We alos evaluated the effectiveness of the pipeline on Whoops! dataset \cite{whoops!}, and included the results in the supp.

%This supports our theory that grasping atypicality is vital for decoding the implicit messages in visual persuasive media.

\setlength{\tabcolsep}{0.8pt}

% To evaluate our pipeline on typical images , we compared LLaVA with Vicuna  on a small-set ARR task. Vicuna ($\mathcal{T}_{\mathcal{V}}$) achieved 71.2\%/48.6\%/33.2\% vs. 66.4\%/42.2\%/28.3\% for LLaVA on prec@1/2/3, demonstrating our approach's effectiveness even on typical images.

% \input{figures/error_distribution_analysis}

\subsection{Further Analysis \& Ablation}
\label{subsec:analysis_ablation}

%Please add the following packages if necessary:
%\usepackage{booktabs, multirow} % for borders and merged ranges
%\usepackage{soul}% for underlines
%\usepackage[table]{xcolor} % for cell colors
%\usepackage{changepage,threeparttable} % for wide tables
%If the table is too wide, replace \begin{table}[!htp]...\end{table} with
%\begin{adjustwidth}{-2.5 cm}{-2.5 cm}\centering\begin{threeparttable}[!htb]...\end{threeparttable}\end{adjustwidth}
\setlength{\tabcolsep}{2.5pt}

\begin{table}[t]%[!htp]
\centering
%We use GPT-4 to create semantically hard negatives.}

\scriptsize
\begin{tabular}{lc||ccc||c}\toprule
\multirow{3}{*}{\textbf{Neg. Strategy}} & \multirow{3}{*}{\textbf{Model}} & \multicolumn{3}{c||}{\textbf{\textit{Multi}}} & \multirow{2}{*}{\textbf{\textit{Single}}} \\
& & \multicolumn{3}{c||}{\textbf{\textit{Precision@k}}} &  \\
&  &\textbf{k=1} &\textbf{k=2} &\textbf{k=3} &\textbf{Acc} \\
\hline
\hline
\multirow{4}{*}{12 Neg.~\cite{hussain2017automatic,jia-etal-2023-kafa}} &CLIP $(I)$ &98.79 &\textbf{97.58} &\textbf{92.20 }&\textbf{96.77} \\
&CLIP $(I+T)$ &97.58 &\textbf{97.58} & 87.10 & 90.32 \\
 % &CLIP $(I + T + V)$ &\textbf{99.60} &\textbf{98.59} &\textbf{93.01} &95.97 \\
 
 &LLaVA $(I)$ &93.47 &74.08 &56.33 & 94.31 \\
 
 % \hline
 
 &GPT4 $(\mathcal{T}_{\mathcal{V}})$ &\textbf{99.60} &96.98 &91.13 &93.52 \\
 
 \hline
 \hline
 
% \multirow{3}{*}{24 Neg. (ASHMIT)}  
%  &LLaVA $(I)$ &76.92 &54.86 &38.60 & \\
%  % &CLIP (or KAFA) & 00.00&00.00 &00.00 & 00.00 \\
%  &GPT4  $(\mathcal{T}_{\mathcal{V}})$ &97.17 &90.89 &77.60 & \\
%  \hline
\multirow{5}{*}{18  Hard Neg.} &CLIP $(I)$ &64.52 &34.48 &22.98 &20.97 \\
&CLIP (I+T) &47.18 &25.40 &16.94 &15.73 \\
 % &CLIP $(I + T + V)$ &70.16 &38.31 &25.54 &29.84 \\
 &LLaVA $(I)$ &59.67 &38.27 &26.06 & 26.80\\
 
 &GPT4 $(\mathcal{T}_{\mathcal{V}})$ &\textbf{96.77} &\textbf{87.30} &\textbf{74.60} &\textbf{96.77} \\
\bottomrule
\end{tabular}
\caption{\textbf{Impact of semantically hard negatives on Small-set.}}
\label{tab:ablation_hard_easy_negative}
\vspace{-5pt}
\end{table}

\textbf{Hard Negatives ablation.}
Table~\ref{tab:ablation_hard_easy_negative} compares our semantically hard negatives against those used in prior work. 
VLM performance drops substantially when faced with our hard negatives, evidenced by 69.22/30.27 drop in CLIP(I)/LLaVA(I) on prec@3. Conversely, our method exhibits a decrease of no more than 17 across all metrics, demonstrating robustness in reasoning compared to VLMs.

% CLIP(I) and LLaVA(I) showing decreases of 69.22 and 30.27 points respectively on prec@3. In comparison, our method's performance declines by no more than 17 points. This demonstrates both the robustness of our approach and the effectiveness of hard negatives in evaluating the reasoning capabilities of VLMs and LLMs.

% VLM performance significantly declines with our hard negatives by 69.22/30.27 in CLIP(I)/LLaVA(I) on prec@3. In contrast, our method's performance  decreases for no more than 17, demonstrating its robustness and effectiveness of hard negatives for evaluating VLMs and LLMs reasoning.

\textbf{Error analysis on ARR.} 
In Fig.~\ref{fig:ARR_Error_Analysis}, VLMs (i.e., LLaVA and CLIP) perform comparably to ($\mathcal{T}_\mathcal{V}$) on `object swap' as these negative options include \textbf{different objects} from the ground-truth, making them easy for VLMs to identify. However, \textbf{VLMs make substantially more errors than GPT-4($\mathcal{T}_\mathcal{V}$)} on semantically incorrect negatives %that are \textbf{semantically different/incorrect} 
(i.e., `action alter,' `reason alter,' `statement alter,' and `adjective alter' ). %For instance, CLIP and LLaVA have over 400 mispredictions on `action alter' (changed/negated action) and `statement alter' (unrelated statement), respectively. 
This confirms VLMs mainly rely on visual elements (e.g., objects) rather than deeper reasoning (examples in supp).

 \textbf{Error analysis on ASR.} In Fig.~\ref{fig:ASR_Error_Analysis}, LLaVA makes notably more errors, particularly on `Wrong Relation' options, which demand deeper reasoning than other negative types.

% \textbf{Verbalization types Ablation}
\textbf{Effectiveness of each component in atypicality-aware verbalization.}
We ablate the effectiveness of each step in atypicality-aware verbalization on the ARR small-set (details in supp).  
The performance of $\mathcal{T}_{\mathcal{V}}$ shows the advantage of atypicality-aware verbalization over basic concatenation ($V+T+IN+UH$). Specifically, $\mathcal{T}_\mathcal{V}$ improves top-1 acc by 14.76  on GPT-3.5 and 12.78 on GPT-4. This is because LLaVA-generated descriptions are inherently noisy, and their naive concatenation can mislead the models. A further issue is increased prompt length. 
Additionally, $UH$ alone is less effective than $IN$, but $\mathcal{T}_\mathcal{V}$ outperforms $\mathcal{T}_\mathcal{V}\setminus UH$, showing that atypicality is important yet complementary to basic image information captured in $V$, $T$, and $IN$.

\textbf{Generalization beyond ads}. We test our atypicality-aware verbalization pipeline on WHOOPS!~\cite{whoops!} in supp.

\section{Conclusion}
\label{sec:conclusion}
This work challenged VLMs on complex rhetorical visual media, focusing on atypicality in advertisements. We introduced three novel atypicality tasks and benchmarked VLMs on and ARR, revealing their limitations in advanced reasoning. Despite these limitations, VLMs showed potential in extracting relevant information for understanding the atypical images. Our atypicality-aware verbalization strategy significantly enhances LLM performance on ARR tasks. Extensive experiments demonstrate that our approach outperforms VLM baselines, proving the effectiveness of incorporating atypicality inference for understanding ad images. These findings highlight the importance of atypicality in interpreting complex visual media and the superior reasoning abilities of LLMs over VLMs.

% ECCV:
% We devised three novel tasks for atypicality understanding, revealing that despite suboptimal performance, VLMs hold potential for extracting pertinent information for atypicality and action-reason analysis. Consequently, 
% we developed an approach for creating atypicality-aware image verbalizations, which an LLM uses to identify the atypicality and action-reason statements. Through extensive experiments, we show our proposed approach outperforms VLM baselines and significantly improves verbalization, proving the effectiveness of atypicality inference for understanding ad images in zero-shot manner. 

\textbf{Limitations.} The PittAd dataset~\cite{hussain2017automatic} is widely used for understanding visual media (e.g., \cite{ye2018advise,akula2023metaclue,jia-etal-2023-kafa,guo2021detecting,mittal2021affect2mm}), but it contains many annotations reflecting human biases and some images with sensitive content. Exploring these biases is beyond the scope of this paper.
%that may potentially be harmful to sensitive characteristics. 
%In this work, we utilized VLMs/LLMs to extract atpyicality and detect the interpretation of the image. Therefore, the performance of our method is impacted by the models used in this paper. However, it is important to acknowledge that understanding atypicality is a non-trivial and complex task, which was not feasible with the traditional CV models.

\textbf{Acknowledgment.} This work was partly supported by NSF Grant No. 2006885. It was also supported by the University of Pittsburgh Center for Research Computing, RRID:SCR\_022735, through the resources provided. Specifically, this work used the H2P cluster, which is supported by NSF award number OAC-2117681. We gratefully acknowledge the support of those who contributed to the human evaluation of this work.

%%%%%%%%% REFERENCES
{\small
\bibliographystyle{ieee_fullname}
\bibliography{egbib}

\begin{thebibliography}{10}\itemsep=-1pt

\bibitem{2023GPT4VisionSC}
Gpt-4v(ision) system card.
\newblock 2023.

\bibitem{akula2023metaclue}
Arjun~R Akula, Brendan Driscoll, Pradyumna Narayana, Soravit Changpinyo, Zhiwei Jia, Suyash Damle, Garima Pruthi, Sugato Basu, Leonidas Guibas, William~T Freeman, et~al.
\newblock Metaclue: Towards comprehensive visual metaphors research.
\newblock In {\em Proceedings of the IEEE/CVF Conference on Computer Vision and Pattern Recognition}, pages 23201--23211, 2023.

\bibitem{alayrac2022flamingo}
Jean-Baptiste Alayrac, Jeff Donahue, Pauline Luc, Antoine Miech, Iain Barr, Yana Hasson, Karel Lenc, Arthur Mensch, Katherine Millican, Malcolm Reynolds, et~al.
\newblock Flamingo: a visual language model for few-shot learning.
\newblock {\em Advances in Neural Information Processing Systems}, 35:23716--23736, 2022.

\bibitem{bavaresco-etal-2024-dont}
Anna Bavaresco, Alberto Testoni, and Raquel Fern{\'a}ndez.
\newblock Don{'}t buy it! {R}eassessing the ad understanding abilities of contrastive multimodal models.
\newblock In {\em ACL (Short)}, Aug. 2024.

\bibitem{bhattacharya2023video}
Aanisha Bhattacharya, Yaman~K Singla, Balaji Krishnamurthy, Rajiv~Ratn Shah, and Changyou Chen.
\newblock A video is worth 4096 tokens: Verbalize story videos to understand them in zero shot.
\newblock In {\em EMNLP}, 2023.

\bibitem{bitton2023vasr}
Yonatan Bitton, Ron Yosef, Eliyahu Strugo, Dafna Shahaf, Roy Schwartz, and Gabriel Stanovsky.
\newblock Vasr: Visual analogies of situation recognition.
\newblock In {\em Proceedings of the AAAI Conference on Artificial Intelligence}, volume~37, pages 241--249, 2023.

\bibitem{whoops!}
Nitzan Bitton-Guetta, Yonatan Bitton, Jack Hessel, Ludwig Schmidt, Yuval Elovici, Gabriel Stanovsky, and Roy Schwartz.
\newblock Breaking common sense: Whoops! a vision-and-language benchmark of synthetic and compositional images.
\newblock In {\em Proceedings of the IEEE/CVF International Conference on Computer Vision}, pages 2616--2627, 2023.

\bibitem{cai2024internlm2}
Zheng Cai, Maosong Cao, Haojiong Chen, Kai Chen, Keyu Chen, Xin Chen, Xun Chen, Zehui Chen, Zhi Chen, Pei Chu, et~al.
\newblock Internlm2 technical report.
\newblock {\em arXiv preprint arXiv:2403.17297}, 2024.

\bibitem{chen2023understanding}
Aochuan Chen, Yuguang Yao, Pin-Yu Chen, Yihua Zhang, and Sijia Liu.
\newblock Understanding and improving visual prompting: A label-mapping perspective.
\newblock In {\em Proceedings of the IEEE/CVF Conference on Computer Vision and Pattern Recognition}, pages 19133--19143, 2023.

\bibitem{chen2015microsoft}
Xinlei Chen, Hao Fang, Tsung-Yi Lin, Ramakrishna Vedantam, Saurabh Gupta, Piotr Doll{\'a}r, and C~Lawrence Zitnick.
\newblock Microsoft coco captions: Data collection and evaluation server.
\newblock {\em arXiv preprint arXiv:1504.00325}, 2015.

\bibitem{Chen_2024_CVPR}
Zhe Chen, Jiannan Wu, Wenhai Wang, Weijie Su, Guo Chen, Sen Xing, Muyan Zhong, Qinglong Zhang, Xizhou Zhu, Lewei Lu, Bin Li, Ping Luo, Tong Lu, Yu Qiao, and Jifeng Dai.
\newblock Internvl: Scaling up vision foundation models and aligning for generic visual-linguistic tasks.
\newblock In {\em Proceedings of the IEEE/CVF Conference on Computer Vision and Pattern Recognition (CVPR)}, pages 24185--24198, June 2024.

\bibitem{vicuna2023}
Wei-Lin Chiang, Zhuohan Li, Zi Lin, Ying Sheng, Zhanghao Wu, Hao Zhang, Lianmin Zheng, Siyuan Zhuang, Yonghao Zhuang, Joseph~E. Gonzalez, Ion Stoica, and Eric~P. Xing.
\newblock Vicuna: An open-source chatbot impressing gpt-4 with 90\%* chatgpt quality, March 2023.

\bibitem{dai2023instructblip}
Wenliang Dai, Junnan Li, Dongxu Li, Anthony Tiong, Junqi Zhao, Weisheng Wang, Boyang Li, Pascale Fung, and Steven Hoi.
\newblock Instruct{BLIP}: Towards general-purpose vision-language models with instruction tuning.
\newblock In {\em Thirty-seventh Conference on Neural Information Processing Systems}, 2023.

\bibitem{dey2021beyond}
Arka~Ujjal Dey, Suman~K Ghosh, Ernest Valveny, and Gaurav Harit.
\newblock Beyond visual semantics: Exploring the role of scene text in image understanding.
\newblock {\em Pattern Recognition Letters}, 149:164--171, 2021.

\bibitem{dziri2024faith}
Nouha Dziri, Ximing Lu, Melanie Sclar, Xiang~Lorraine Li, Liwei Jiang, Bill~Yuchen Lin, Sean Welleck, Peter West, Chandra Bhagavatula, Ronan Le~Bras, et~al.
\newblock Faith and fate: Limits of transformers on compositionality.
\newblock {\em Advances in Neural Information Processing Systems}, 36, 2024.

\bibitem{ge2023improving}
Yunhao Ge, Jie Ren, Andrew Gallagher, Yuxiao Wang, Ming-Hsuan Yang, Hartwig Adam, Laurent Itti, Balaji Lakshminarayanan, and Jiaping Zhao.
\newblock Improving zero-shot generalization and robustness of multi-modal models.
\newblock In {\em Proceedings of the IEEE/CVF conference on computer vision and pattern recognition}, pages 11093--11101, 2023.

\bibitem{goyal2017making}
Yash Goyal, Tejas Khot, Douglas Summers-Stay, Dhruv Batra, and Devi Parikh.
\newblock Making the v in vqa matter: Elevating the role of image understanding in visual question answering.
\newblock In {\em Proceedings of the IEEE conference on computer vision and pattern recognition}, pages 6904--6913, 2017.

\bibitem{guo2021detecting}
Meiqi Guo, Rebecca Hwa, and Adriana Kovashka.
\newblock Detecting persuasive atypicality by modeling contextual compatibility.
\newblock In {\em Proceedings of the IEEE/CVF International Conference on Computer Vision}, pages 972--982, 2021.

\bibitem{hussain2017automatic}
Zaeem Hussain, Mingda Zhang, Xiaozhong Zhang, Keren Ye, Christopher Thomas, Zuha Agha, Nathan Ong, and Adriana Kovashka.
\newblock Automatic understanding of image and video advertisements.
\newblock In {\em Proceedings of the IEEE conference on computer vision and pattern recognition}, pages 1705--1715, 2017.

\bibitem{jia-etal-2023-kafa}
Zhiwei Jia, Pradyumna Narayana, Arjun Akula, Garima Pruthi, Hao Su, Sugato Basu, and Varun Jampani.
\newblock {KAFA}: Rethinking image ad understanding with knowledge-augmented feature adaptation of vision-language models.
\newblock In Sunayana Sitaram, Beata Beigman~Klebanov, and Jason~D Williams, editors, {\em Proceedings of the 61st Annual Meeting of the Association for Computational Linguistics (Volume 5: Industry Track)}, pages 772--785, Toronto, Canada, July 2023. Association for Computational Linguistics.

\bibitem{kalra2020understanding}
Kanika Kalra, Bhargav Kurma, Silpa~Vadakkeeveetil Sreelatha, Manasi Patwardhan, and Shirish Karande.
\newblock Understanding advertisements with bert.
\newblock In {\em Proceedings of the 58th Annual Meeting of the Association for Computational Linguistics}, pages 7542--7547, 2020.

\bibitem{khattak2023maple}
Muhammad~Uzair Khattak, Hanoona Rasheed, Muhammad Maaz, Salman Khan, and Fahad~Shahbaz Khan.
\newblock Maple: Multi-modal prompt learning.
\newblock In {\em Proceedings of the IEEE/CVF Conference on Computer Vision and Pattern Recognition}, pages 19113--19122, 2023.

\bibitem{kojima2022large}
Takeshi Kojima, Shixiang~Shane Gu, Machel Reid, Yutaka Matsuo, and Yusuke Iwasawa.
\newblock Large language models are zero-shot reasoners.
\newblock {\em Advances in neural information processing systems}, 35:22199--22213, 2022.

\bibitem{lan2023improving}
Yunshi Lan, Xiang Li, Xin Liu, Yang Li, Wei Qin, and Weining Qian.
\newblock Improving zero-shot visual question answering via large language models with reasoning question prompts.
\newblock In {\em Proceedings of the 31st ACM International Conference on Multimedia}, pages 4389--4400, 2023.

\bibitem{li2023blip}
Junnan Li, Dongxu Li, Silvio Savarese, and Steven Hoi.
\newblock Blip-2: Bootstrapping language-image pre-training with frozen image encoders and large language models.
\newblock In {\em ICML}, 2023.

\bibitem{NEURIPS2023_761c3284}
Liunian Li, Zi-Yi Dou, Nanyun Peng, and Kai-Wei Chang.
\newblock Desco: Learning object recognition with rich language descriptions.
\newblock In A. Oh, T. Neumann, A. Globerson, K. Saenko, M. Hardt, and S. Levine, editors, {\em Advances in Neural Information Processing Systems}, volume~36, pages 37511--37526. Curran Associates, Inc., 2023.

\bibitem{li-etal-2024-enhancing-advanced}
Zhiyuan Li, Dongnan Liu, Chaoyi Zhang, Heng Wang, Tengfei Xue, and Weidong Cai.
\newblock Enhancing advanced visual reasoning ability of large language models.
\newblock In Yaser Al-Onaizan, Mohit Bansal, and Yun-Nung Chen, editors, {\em Proceedings of the 2024 Conference on Empirical Methods in Natural Language Processing}, pages 1915--1929, Miami, Florida, USA, Nov. 2024. Association for Computational Linguistics.

\bibitem{liu2023improvedllava}
Haotian Liu, Chunyuan Li, Yuheng Li, and Yong~Jae Lee.
\newblock Improved baselines with visual instruction tuning.
\newblock In {\em Proceedings of the IEEE/CVF Conference on Computer Vision and Pattern Recognition}, pages 26296--26306, 2024.

\bibitem{liu2023visual}
Haotian Liu, Chunyuan Li, Qingyang Wu, and Yong~Jae Lee.
\newblock Visual instruction tuning.
\newblock In {\em NeurIPS}, 2023.

\bibitem{lu2024chameleon}
Pan Lu, Baolin Peng, Hao Cheng, Michel Galley, Kai-Wei Chang, Ying~Nian Wu, Song-Chun Zhu, and Jianfeng Gao.
\newblock Chameleon: Plug-and-play compositional reasoning with large language models.
\newblock {\em Advances in Neural Information Processing Systems}, 36, 2024.

\bibitem{mcquarrie1999visual}
Edward~F McQuarrie and David~Glen Mick.
\newblock Visual rhetoric in advertising: Text-interpretive, experimental, and reader-response analyses.
\newblock {\em Journal of consumer research}, 26(1):37--54, 1999.

\bibitem{mitra2024compositional}
Chancharik Mitra, Brandon Huang, Trevor Darrell, and Roei Herzig.
\newblock Compositional chain-of-thought prompting for large multimodal models.
\newblock In {\em Proceedings of the IEEE/CVF Conference on Computer Vision and Pattern Recognition}, pages 14420--14431, 2024.

\bibitem{mittal2021affect2mm}
Trisha Mittal, Puneet Mathur, Aniket Bera, and Dinesh Manocha.
\newblock Affect2mm: Affective analysis of multimedia content using emotion causality.
\newblock In {\em Proceedings of the IEEE/CVF Conference on Computer Vision and Pattern Recognition}, pages 5661--5671, 2021.

\bibitem{mokady2021clipcap}
Ron Mokady, Amir Hertz, and Amit~H Bermano.
\newblock Clipcap: Clip prefix for image captioning.
\newblock {\em arXiv preprint arXiv:2111.09734}, 2021.

\bibitem{radford2021learning}
Alec Radford, Jong~Wook Kim, Chris Hallacy, Aditya Ramesh, Gabriel Goh, Sandhini Agarwal, Girish Sastry, Amanda Askell, Pamela Mishkin, Jack Clark, et~al.
\newblock Learning transferable visual models from natural language supervision.
\newblock In {\em International conference on machine learning}, pages 8748--8763. PMLR, 2021.

\bibitem{reinartz2013creativity}
Werner Reinartz and Peter Saffert.
\newblock Creativity in advertising: When it works and when it doesn’t.
\newblock {\em Harvard Business Review}, 91(6):106--111, 2013.

\bibitem{safaei2024active}
Bardia Safaei and Vishal~M Patel.
\newblock Active learning for vision-language models.
\newblock {\em arXiv preprint arXiv:2410.22187}, 2024.

\bibitem{smith2007modeling}
Robert~E Smith, Scott~B MacKenzie, Xiaojing Yang, Laura~M Buchholz, and William~K Darley.
\newblock Modeling the determinants and effects of creativity in advertising.
\newblock {\em Marketing science}, 26(6):819--833, 2007.

\bibitem{song2020mpnet}
Kaitao Song, Xu Tan, Tao Qin, Jianfeng Lu, and Tie-Yan Liu.
\newblock Mpnet: Masked and permuted pre-training for language understanding.
\newblock {\em Advances in neural information processing systems}, 33:16857--16867, 2020.

\bibitem{thrush2022winoground}
Tristan Thrush, Ryan Jiang, Max Bartolo, Amanpreet Singh, Adina Williams, Douwe Kiela, and Candace Ross.
\newblock Winoground: Probing vision and language models for visio-linguistic compositionality.
\newblock In {\em Proceedings of the IEEE/CVF Conference on Computer Vision and Pattern Recognition}, pages 5238--5248, 2022.

\bibitem{Touvron2023Llama2O}
Hugo Touvron, Louis Martin, Kevin~R. Stone, Peter Albert, Amjad Almahairi, Yasmine Babaei, Nikolay Bashlykov, Soumya Batra, Prajjwal Bhargava, Shruti Bhosale, Daniel~M. Bikel, Lukas Blecher, Cristian~Cant{\'o}n Ferrer, Moya Chen, Guillem Cucurull, David Esiobu, Jude Fernandes, Jeremy Fu, Wenyin Fu, Brian Fuller, Cynthia Gao, Vedanuj Goswami, Naman Goyal, Anthony~S. Hartshorn, Saghar Hosseini, Rui Hou, Hakan Inan, Marcin Kardas, Viktor Kerkez, Madian Khabsa, Isabel~M. Kloumann, A.~V. Korenev, Punit~Singh Koura, Marie-Anne Lachaux, Thibaut Lavril, Jenya Lee, Diana Liskovich, Yinghai Lu, Yuning Mao, Xavier Martinet, Todor Mihaylov, Pushkar Mishra, Igor Molybog, Yixin Nie, Andrew Poulton, Jeremy Reizenstein, Rashi Rungta, Kalyan Saladi, Alan Schelten, Ruan Silva, Eric~Michael Smith, R. Subramanian, Xia Tan, Binh Tang, Ross Taylor, Adina Williams, Jian~Xiang Kuan, Puxin Xu, Zhengxu Yan, Iliyan Zarov, Yuchen Zhang, Angela Fan, Melanie Kambadur, Sharan Narang, Aurelien Rodriguez, Robert Stojnic, Sergey Edunov, and
  Thomas Scialom.
\newblock Llama 2: Open foundation and fine-tuned chat models.
\newblock {\em ArXiv}, abs/2307.09288, 2023.

\bibitem{tsimpoukelli2021multimodal}
Maria Tsimpoukelli, Jacob~L Menick, Serkan Cabi, SM Eslami, Oriol Vinyals, and Felix Hill.
\newblock Multimodal few-shot learning with frozen language models.
\newblock {\em Advances in Neural Information Processing Systems}, 34:200--212, 2021.

\bibitem{vedantam2015learning}
Ramakrishna Vedantam, Xiao Lin, Tanmay Batra, C~Lawrence Zitnick, and Devi Parikh.
\newblock Learning common sense through visual abstraction.
\newblock In {\em Proceedings of the IEEE international conference on computer vision}, pages 2542--2550, 2015.

\bibitem{wang2024t}
Lei Wang, Yi Hu, Jiabang He, Xing Xu, Ning Liu, Hui Liu, and Heng~Tao Shen.
\newblock T-sciq: Teaching multimodal chain-of-thought reasoning via large language model signals for science question answering.
\newblock In {\em Proceedings of the AAAI Conference on Artificial Intelligence}, volume~38, pages 19162--19170, 2024.

\bibitem{wei2022chain}
Jason Wei, Xuezhi Wang, Dale Schuurmans, Maarten Bosma, Fei Xia, Ed Chi, Quoc~V Le, Denny Zhou, et~al.
\newblock Chain-of-thought prompting elicits reasoning in large language models.
\newblock {\em Advances in Neural Information Processing Systems}, 35:24824--24837, 2022.

\bibitem{williamson1978decoding}
Judith Williamson.
\newblock {\em Decoding advertisements}, volume~4.
\newblock Marion Boyars London, 1978.

\bibitem{ye2018advise}
Keren Ye and Adriana Kovashka.
\newblock Advise: Symbolism and external knowledge for decoding advertisements.
\newblock In {\em Proceedings of the European Conference on Computer Vision (ECCV)}, pages 837--855, 2018.

\bibitem{yeinterpreting}
Keren Ye, Narges~Honarvar Nazari, James Hahn, Zaeem Hussain, Mingda Zhang, and Adriana Kovashka.
\newblock Interpreting the rhetoric of visual advertisements.
\newblock {\em IEEE transactions on pattern analysis and machine intelligence}, 43(4):1308--1323, 2019.

\bibitem{Ye_2021_WACV}
Keren Ye, Mingda Zhang, and Adriana Kovashka.
\newblock Breaking shortcuts by masking for robust visual reasoning.
\newblock In {\em Proceedings of the IEEE/CVF Winter Conference on Applications of Computer Vision (WACV)}, pages 3520--3530, January 2021.

\bibitem{you2023idealgpt}
Haoxuan You, Rui Sun, Zhecan Wang, Long Chen, Gengyu Wang, Hammad Ayyubi, Kai-Wei Chang, and Shih-Fu Chang.
\newblock Idealgpt: Iteratively decomposing vision and language reasoning via large language models.
\newblock In {\em Findings of the Association for Computational Linguistics: EMNLP 2023}, pages 11289--11303, 2023.

\bibitem{yuksekgonul2023when}
Mert Yuksekgonul, Federico Bianchi, Pratyusha Kalluri, Dan Jurafsky, and James Zou.
\newblock When and why vision-language models behave like bags-of-words, and what to do about it?
\newblock In {\em International Conference on Learning Representations}, 2023.

\bibitem{zellers2019recognition}
Rowan Zellers, Yonatan Bisk, Ali Farhadi, and Yejin Choi.
\newblock From recognition to cognition: Visual commonsense reasoning.
\newblock In {\em Proceedings of the IEEE/CVF conference on computer vision and pattern recognition}, pages 6720--6731, 2019.

\bibitem{zhang2023multimodal}
Zhuosheng Zhang, Aston Zhang, Mu Li, George Karypis, Alex Smola, et~al.
\newblock Multimodal chain-of-thought reasoning in language models.
\newblock {\em Transactions on Machine Learning Research}.

\bibitem{zheng2023ddcot}
Ge Zheng, Bin Yang, Jiajin Tang, Hong-Yu Zhou, and Sibei Yang.
\newblock Ddcot: Duty-distinct chain-of-thought prompting for multimodal reasoning in language models.
\newblock {\em Advances in Neural Information Processing Systems}, 36:5168--5191, 2023.

\bibitem{zhou2023rome}
Kankan Zhou, Eason Lai, Wei Bin~Au Yeong, Kyriakos Mouratidis, and Jing Jiang.
\newblock Rome: Evaluating pre-trained vision-language models on reasoning beyond visual common sense.
\newblock In {\em Findings of the Association for Computational Linguistics: EMNLP 2023}, pages 10185--10197, 2023.

\bibitem{zhou2022conditional}
Kaiyang Zhou, Jingkang Yang, Chen~Change Loy, and Ziwei Liu.
\newblock Conditional prompt learning for vision-language models.
\newblock In {\em Proceedings of the IEEE/CVF Conference on Computer Vision and Pattern Recognition}, pages 16816--16825, 2022.

\bibitem{zhou2022learning}
Kaiyang Zhou, Jingkang Yang, Chen~Change Loy, and Ziwei Liu.
\newblock Learning to prompt for vision-language models.
\newblock {\em International Journal of Computer Vision}, 130(9):2337--2348, 2022.

\bibitem{zhu2024minigpt}
Deyao Zhu, Jun Chen, Xiaoqian Shen, Xiang Li, and Mohamed Elhoseiny.
\newblock Mini{GPT}-4: Enhancing vision-language understanding with advanced large language models.
\newblock In {\em The Twelfth International Conference on Learning Representations}, 2024.

\end{thebibliography}
}
\clearpage

\section{Supplement}
We aimed to investigate the effectiveness of VLMs for understanding persuasive advertisements. Concretely, we hypothesized that understanding atypicality can aid understanding advertisements. Hence, we first compared state-of-the-art VLMs on three novel atypicality understanding tasks: (1) Multi-label Atypicality Classification (MAC), (2) Atypicality Statement Retrieval (ASR), and (3) Atypicality Object Recognition (AOR). Table~\ref{tab:supp_MAC_fullset_full_metrics} compares the performance of VLMs with our proposed strategies on the MAC task, offering a more comprehensive evaluation of metrics than Table 1 in the main paper. Table~\ref{tab:atyp-obj-recog-small} summarizes the results on the small-set for the AOR task. Full-set results on ASR and AOR tasks can be found in Table 1 and Table 2 of the main paper.

\begin{figure*}[!tp]
    \scriptsize
    \centering
    \begin{subfigure}[b]{0.4\textwidth}
        \includegraphics[width=\textwidth]{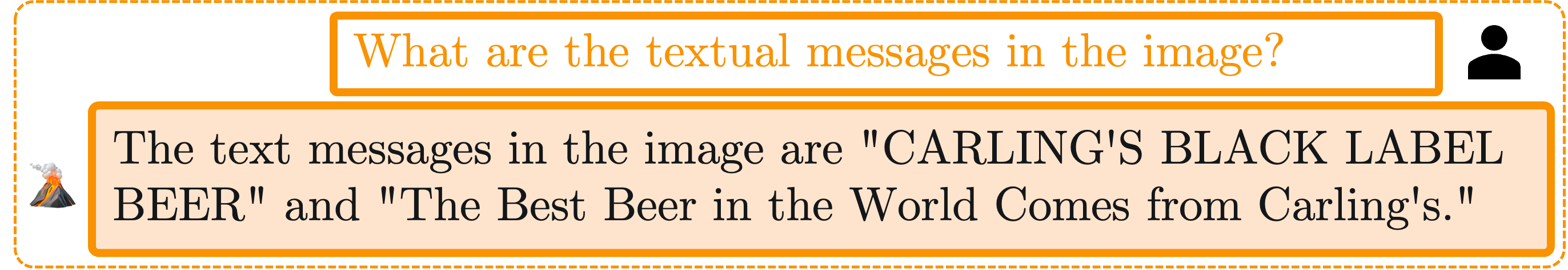}
        \caption{Scene-text ($T$) detection}
        \label{fig:full_pipeline_sub1}
    \end{subfigure}
    \hfill % This adds space between the subfigures
    \begin{subfigure}[b]{0.4\textwidth}
        \includegraphics[width=\textwidth]{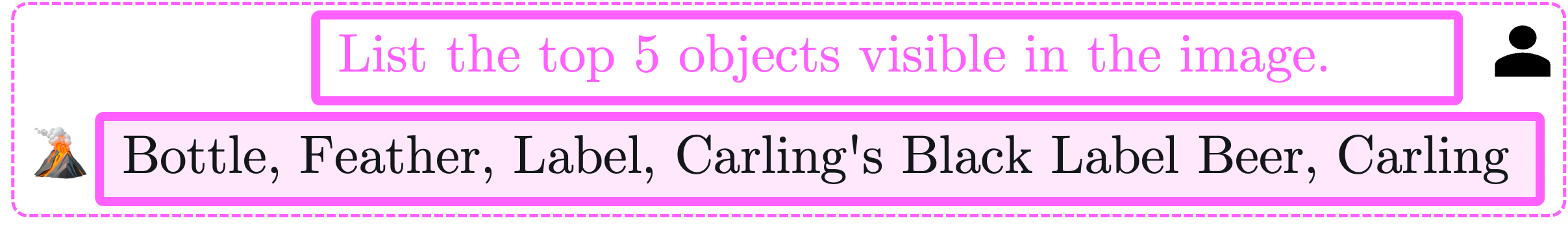}
        \caption{Top-5 objects ($V$) detection}
        \label{fig:full_pipeline_sub2}
    \end{subfigure}
    \hfill % This adds space between the subfigures
    \begin{subfigure}[b]{0.8\textwidth}
        \includegraphics[width=\textwidth]{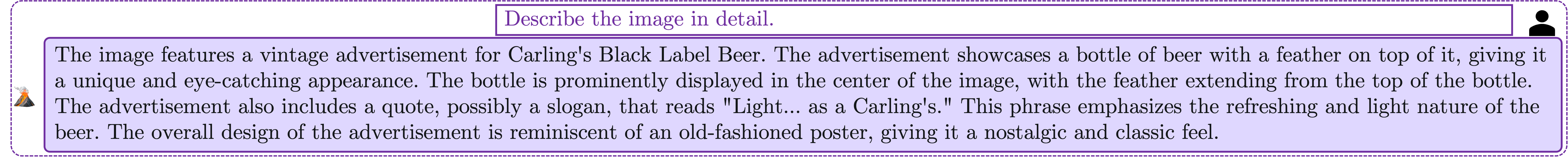}
        \caption{ ImageNarrator ($IN$) description generation}
        \label{fig:full_pipeline_sub3}
    \end{subfigure}

    \begin{subfigure}[b]{0.8\textwidth}
        \includegraphics[width=\textwidth]{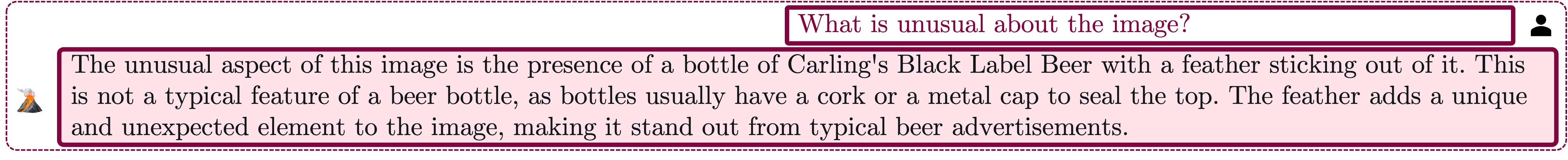}
        \caption{UnusualHighlighter ($UH$) description generation}
        \label{fig:full_pipeline_sub4}
    \end{subfigure}
    \hfill
    \begin{subfigure}[b]{0.8\textwidth}
        \includegraphics[width=\textwidth]{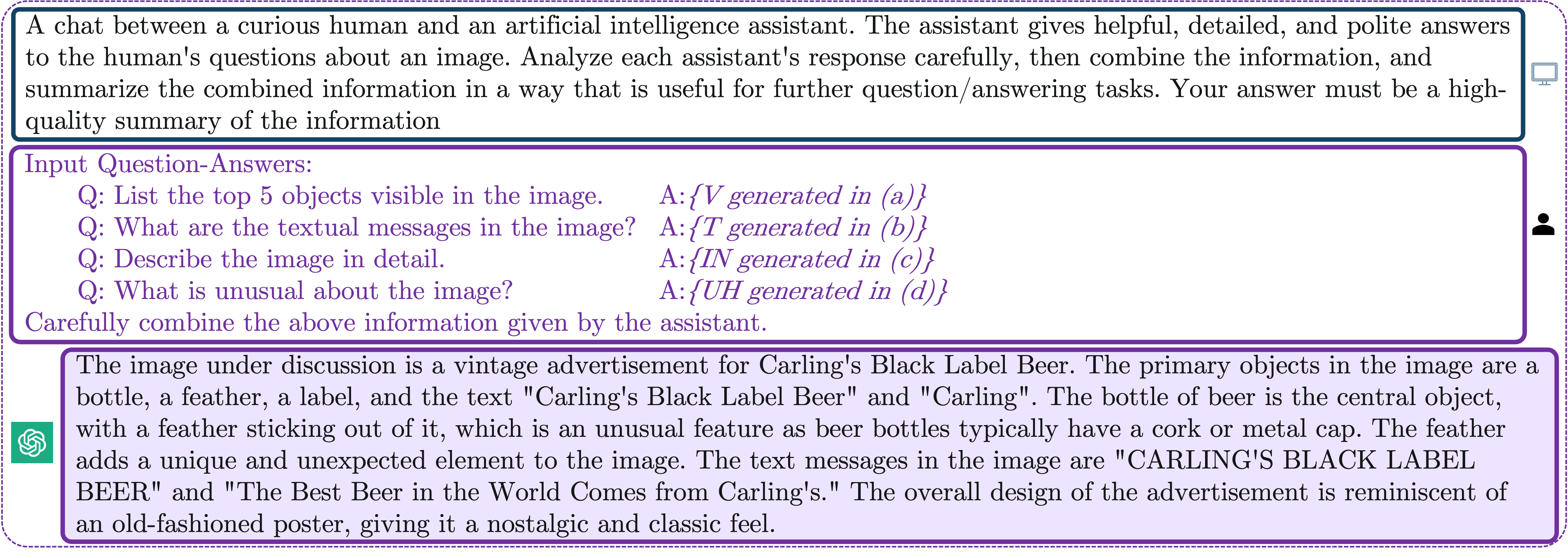}
        \caption{Image description combination ($\mathcal{T}_\mathcal{V}$)}
        \label{fig:full_pipeline_sub5}
    \end{subfigure}
    \hfill
    \begin{subfigure}[b]{0.8\textwidth}
        \includegraphics[width=\textwidth]{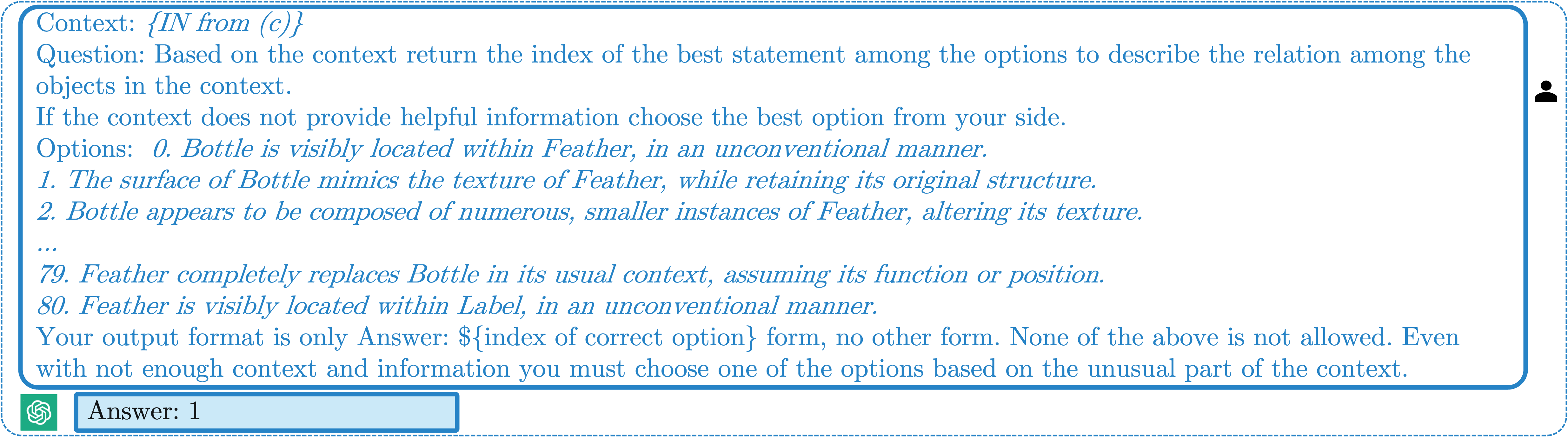}
        \caption{Atyipcality statement retrieval ($\hat{s}$)}
        \label{fig:full_pipeline_sub6}
    \end{subfigure}

    \begin{subfigure}[b]{0.8\textwidth}
        \centering
        \includegraphics[width=\textwidth]{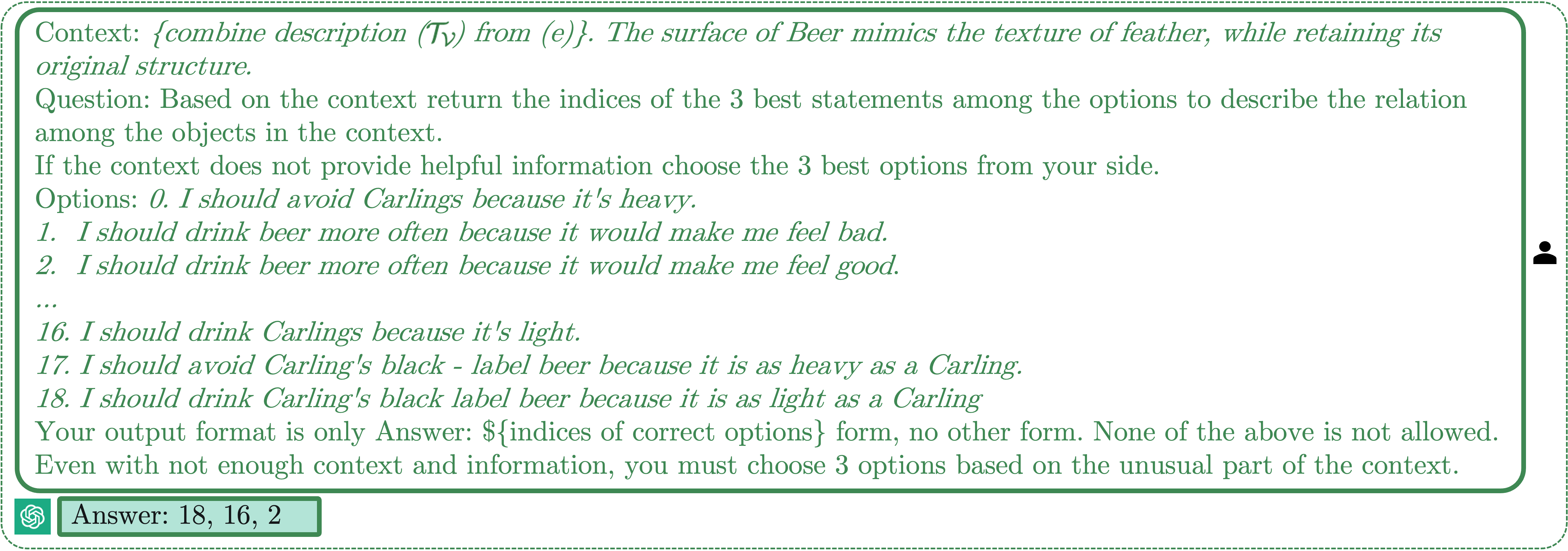}
        \caption{multi-ARR}
        \label{fig:full_pipeline_sub7}
    \end{subfigure}
    
    \caption{\textbf{Full pipeline for the multi-ARR task}. (a-d) Image verbalization with LLaVA, (e) Outputs of (a-d) are input into GPT-4 to generate the combined description $\mathcal{T}_\mathcal{V}$, (f) $V$ and atypicality statement templates $\mathcal{S}_{\mathcal{A}}$ generate atypicality statement options. Next, we use $IN$ to retrieve the atypicality statement $\hat{s}$. (g) Finally, we concatenate $\hat{s}$ with $\mathcal{T}_\mathcal{V}$ for mulit-ARR. \{\}/\textit{italic} denote variable/dynamic information.}
    \label{fig:full_pipeline}
\end{figure*}

Secondly, to evaluate the impact of atypicality in ad understanding and analyze VLMs' reasoning ability about atypicality, we proposed a novel atypicality-aware verbalization method. We compared our method with VLMs and verbalization baselines (i.e., $V+T$). Table~\ref{tab:clip_ablation} compares various methods of constructing atypicality-aware verbalization, including concatenation and LLM-based combinations when used with CLIP. We also benchmark these against the CLIP ($I$) baseline and a related zero-shot for KAFA (CLIP ($I + T$)). Full-set results on ARR are in Table 3 in the main paper. Table~\ref{tab:tab_arr_small} ablate different types of verbalization and shows the effectiveness of each component in our proposed verbalization method, which is discussed in Sec.\ref{sec:arr_supp} and Sec.5.3 in the main paper. Table~\ref{tab:typical_multi_selection_action_reason} shows the evaluation of the our method's generalization to the typical images. In Table~\ref{tab:whoops}, we evaluated LLaVA and our method on WHOOPS!~\cite{whoops!}. We further provide analysis for validating the generated semantically hard negatives by GPT-4 in Sec.~\ref{sec:arr_supp} (analysis are in text). An example of our full pipeline for multi action-reason retrieval tasks is demonstrated in Fig.~\ref{fig:full_pipeline}. 

\begin{figure*}[!tp]
    \centering
    \includegraphics[width=1\textwidth]{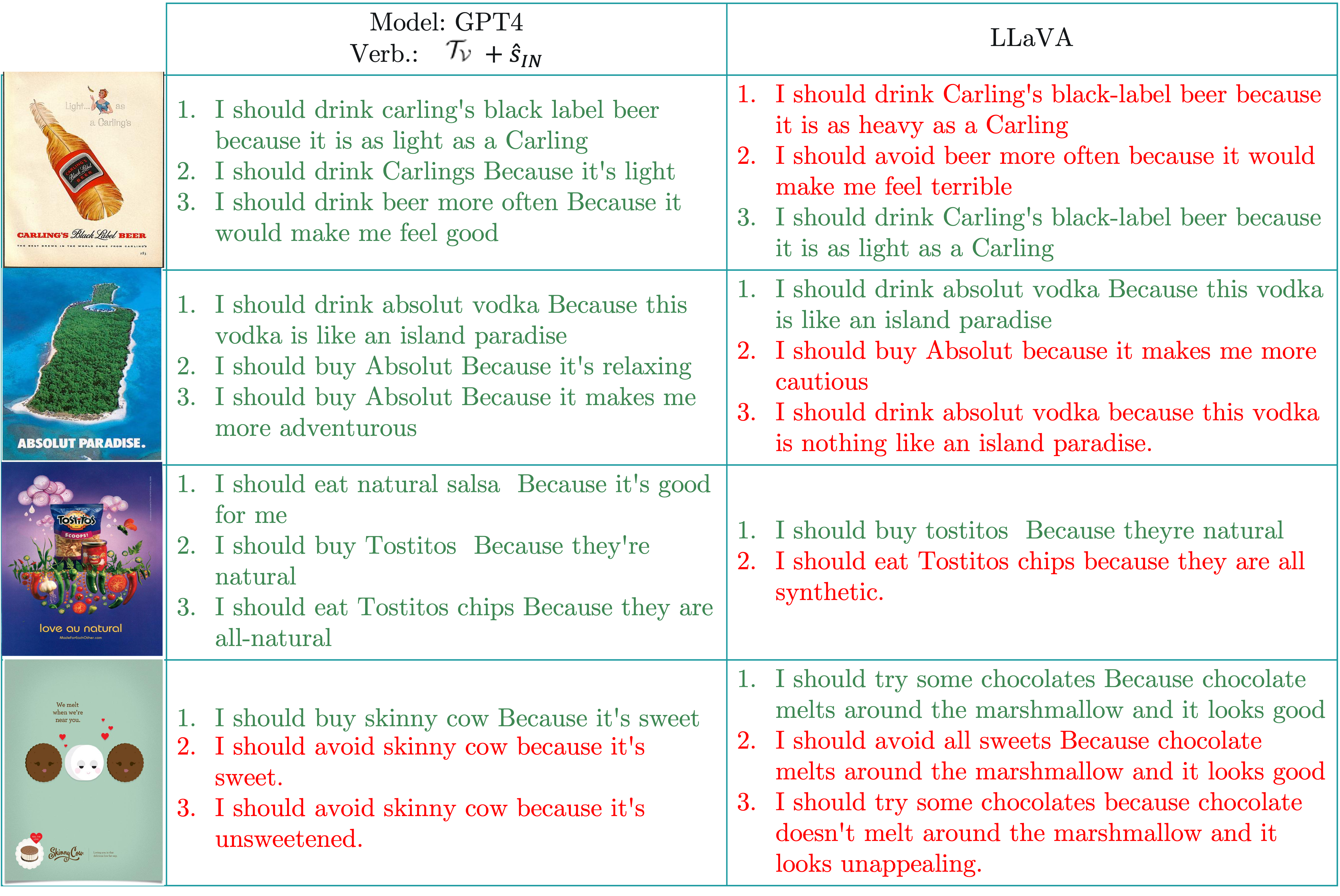}
    \caption{\textbf{Examples of output from Ours (i.e., GPT-4 ($\mathcal{T}_\mathcal{V} + \hat{s}_{IN}$)) and LLaVA in the multi-ARR task.} \textcolor{Green}{Green}/\textcolor{red}{Red} denote correct/incorrect predictions, respectively. 
    }
    \label{fig:comparison_example}
\end{figure*}

Figs.~\ref{fig:HN_examples} and \ref{fig:comparison_example} visualize examples of semantically hard negatives and a comparison between the predictions of our proposed method and LLaVA, respectively. Finally, the prompts utilized in this study are detailed in Sec.~\ref{supp:prompts}.

% In Sec.~\ref{supp:prompts} we provide the prompts used for each step in our method, and the baselines for different tasks.

% \section{Results}

\subsection{Atypicality Understanding Results}
\label{supp:atypicality_understanding_tasks}
%Please add the following packages if necessary:
%\usepackage{booktabs, multirow} % for borders and merged ranges
%\usepackage{soul}% for underlines
%\usepackage[table]{xcolor} % for cell colors
%\usepackage{changepage,threeparttable} % for wide tables
%If the table is too wide, replace \begin{table}[!htp]...\end{table} with
%\begin{adjustwidth}{-2.5 cm}{-2.5 cm}\centering\begin{threeparttable}[!htb]...\end{threeparttable}\end{adjustwidth}
% \setlength{\tabcolsep}{2pt}
\setlength{\tabcolsep}{2pt}
\begin{table}[!tp]
\centering
\scriptsize
\begin{tabular}{lc||cc|cc|cc}\toprule
\multirow{2}{*}{\textbf{Model}} &\multirow{2}{*}{\textbf{Verb.}} &\multicolumn{2}{c}{\textbf{AUC-ROC}} &\multicolumn{2}{c}{\textbf{AUC-PR}} &\multicolumn{2}{c}{\textbf{Subset-Acc}} \\

& & $\checkmark$ & $\times$ & $\checkmark$ & $\times$ & $\checkmark$ & $\times$ \\

\hline
\hline

LLaVA &- &50.16 &50.12 &35.81 &30.46 &0.94 &2.83 \\
InsturctBLIP &- &50.13 &50.03 &35.81 &30.44 &0.51 &1.54 \\

\hline
\hline

\multirow{3}{*}{Vicuna} &$T+V$ &50.63 &50.31 &36.03 &30.53 &3.51 &6.68 \\

&$IN$ &\textbf{52.26} &52.25 &\textbf{36.84} &\textbf{31.50} &4.28 &7.88 \\

&$UH$ &52.03 &\textbf{52.26} &36.64 &31.40 &\textbf{5.22} &\textbf{10.70} \\

\hline

\multirow{3}{*}{GPT-3.5} &$T+V$ &52.40 &51.83 &37.01 &31.34 &\textbf{10.10} &\textbf{24.32} \\

&$IN$ &53.28 &52.71 &37.69 &32.13 &4.20 &9.08 \\

&$UH$ &\textbf{54.36} &\textbf{54.64} &\textbf{38.34} &\textbf{33.17} &7.62 &20.89 \\

\hline

\multirow{3}{*}{GPT 4} &$T+V$ &51.10 &50.91 &36.34 &30.94 &1.71 &3.68 \\

&$IN$ &54.13 &53.88 &38.46 &33.16 &4.79 &9.85 \\

&$UH$ &\textbf{55.51} &\textbf{56.00} &\textbf{39.32} &\textbf{34.41} &\textbf{11.22} &\textbf{28.00} \\
\bottomrule
\end{tabular}
\caption{\textbf{Multi-label atypicality classification on Full-set.} $\checkmark$/$\times$ denotes performance with/without No Atypicality (NA) class. \textbf{Bolded} numbers indicate best-performing strategy per LLM.}\label{tab:supp_MAC_fullset_full_metrics}
\end{table}

\begin{adjustwidth}{-4.5 cm}{-4.5 cm}\centering
\begin{table}[!tp]
\centering
\scriptsize
\begin{tabular}{l<{\hspace{0.25em}}||>{\hspace{0.25em}}c|>{\hspace{0.5em}}c>{\hspace{1em}}c>{\hspace{1em}}c>{\hspace{1em}}c}

\toprule
\multirow{2}{*}{\textbf{Model}} & {\textbf{\textit{Avg. }similarity}} & \multicolumn{3}{c}{\textbf{\textit{\% of scores}}} \\
& \textbf{score} & \textbf{$>$ 0.7} & \textbf{$>$ 0.6}  & \textbf{$>$ 0.5 } \\
\hline
\hline
BLIP2 \cite{li2023blip} & 0.45 & 8.13 & 19.11 & 36.59 \\
InstructBLIP \cite{dai2023instructblip} & 0.47 & 10.57 & 23.58 & 43.90 \\
MiniGPT4 \cite{zhu2024minigpt} & 0.52 & 15.45 & 31.71 & 56.50 \\
LLaVA \cite{liu2023visual} & 0.60 & 29.79 & 56.17 & 69.36 \\
GPT-4V \cite{2023GPT4VisionSC} & \textbf{0.67} & \textbf{46.94} & \textbf{61.63} & \textbf{77.14} \\
\bottomrule
\end{tabular}
\caption{\textbf{Atypical Object Recognition (AOR) on Small-set}. MPNet sentence similarity scores and score thresholds are reported.}
\label{tab:atyp-obj-recog-small}
\end{table}
\end{adjustwidth}
\textbf{Multi-label Atypicality Classification and Atypicality Statement Retrieval.}
Table~\ref{tab:supp_MAC_fullset_full_metrics} presents additional evaluation metrics (i.e., AUC-ROC, AUC-PR, and Subset-Acc) compared to the main paper's table. We observe that $UH$ consistently outperforms all other strategies in AUC-ROC when excluding NA (denoted as $\times$). Similarly, in most LLMs, $UH$ surpasses both $IN$ and $V+T$, showcasing the effectiveness of $UH$ in highlighting the unusualness of the image to facilitate understanding of atypicality. A significant difference is observed between the performance of LLMs on $UH$ and VLMs on subset-acc. Subset-acc is a challenging metric where a prediction is considered correct only if it can successfully identify all atypicalities of the image. For instance, $UH$ on GPT-4 achieves 28\% accuracy, improving LLaVA and InstructBLIP by 25.17 and 26.46 percent, respectively. This underscores the limitations of VLMs in directly recognizing atypicality.

\textbf{Atypical Object Retrieval.}
Table~\ref{tab:atyp-obj-recog-small}  compares current state-of-the-art VLMs on the Atypical Object Recognition (AOR) task. GPT-4V achieves the avg. similarity score of 0.67 between generated statement $\hat{s}=(a^+, \hat{o}^p, \hat{o}^s)$ and ground-truth statement $s=(a^+, o^{+p}, o^{+s})$ with 46.94\% of the scores above 0.7. This is significantly higher than public LLMs, led by LLaVA, where only 29.79 scores are higher than 0.7. While these results show that GPT-4V is more powerful than public VLMs, it is still limited in accurately recognizing the first/second objects and the atypical relationship among them. 
\subsection{Action-Reason Retrieval Results}
\label{sec:arr_supp}
\begin{table}
\centering
%is based on $UH$. LLaVA + GPT-4 verbalization. 
%, $\hat{s}_{UH}$: atypc prediction based on UH}
\vspace{-0.25cm}
\scriptsize
\begin{tabular}{lc||c|c|c|c}
\toprule
 \textbf{Classifier} & \textbf{Verb}. & \multicolumn{3}{c}{\textit{\textbf{Precision@k}}} \\
&&\textbf{k=1} & \textbf{k=2} &\textbf{k=3} & \textbf{avg}\\
\hline
\hline
 \multirow{2}{*}{LLaVA} 
 &$I$  & 59.67 & 38.27  & 26.06 & 41.33\\
 &$I$ ($CoT$) & 66.53 & 42.24 & 28.29 & 45.68 \\
 % \hline
\multirow{1}{*}{Vicuna} 
&$\mathcal{T}_\mathcal{V} + \hat{s}_{IN}$ (Ours) & \textbf{71.77} & \textbf{46.77} & \textbf{31.59} & \textbf{50.04}\\
\hline
\multirow{2}{*}{GPT-4} 
&$\mathcal{T}_\mathcal{V} + \hat{s}_{IN}$ (Ours) & \textbf{96.77}& \textbf{87.77} & \textbf{73.65} & \textbf{86.06}\\
&$\mathcal{T}_\mathcal{V} + \hat{s}_{IN} (CoT)$ & 95.97 & 86.29 & 72.17 & 84.81 \\
\bottomrule
\end{tabular}
\vspace{-0.2cm}
\caption{\textbf{Chain-of-thought prompting %for LLaVA \& GPT-4}, 
for ARR on Small-set}}
\label{tab:CoT}
\vspace{-4pt}
\end{table}

\textbf{Comparison against Chain-of-Thought.} Table~\ref{tab:CoT} compares our proposed atypicality-aware verbalization against CoT (`think step-by-step') in \cite{kojima2022large}. While CoT reasoning yields marginal improvements in LLaVA due to the multi-step nature of the problem, it still falls short when compared to our approach, with significant differences of 4.36 in Vicuna and 40.38 in GPT-4.

We also observed that applying CoT on top of our method in GPT -4 results in lower performance. This happens because our approach already includes a form of implicit reasoning similar to CoT. Adding explicit CoT reasoning creates redundancy, which complicates the reasoning process and may introduce unnecessary steps. This overlap leads to the performance drop, as the extra reasoning adds complexity without improving results.

\begin{table}
\centering

%is based on $UH$. LLaVA + GPT-4 verbalization. 
%, $\hat{s}_{UH}$: atypc prediction based on UH}
\scriptsize
\begin{tabular}{lc||c|c|c|c}
\toprule
 \textbf{Classifier} & \textbf{Verb}. & \multicolumn{3}{c}{\textit{\textbf{Precision@k}}} & \\

&&\textbf{k=1} & \textbf{k=2} &\textbf{k=3} & \textbf{avg}\\
\hline
\hline
 \multirow{1}{*}{LLaVA 1.6} 
 &$I$  & 74.79 & 52.00 &35.73 & 54.17 \\
 % \multirow{1}{*}{InternVL2‑8B} &$I$  & 92.12 & 76.66 & 56.19\\
 % \hline
\multirow{1}{*}{Vicuna} 
&$\mathcal{T}_\mathcal{V}$ (Ours) & \textbf{86.40} & \textbf{62.40} & \textbf{43.19}  & \textbf{63.99}\\
\hline
% \multirow{1}{*}{InternLM} 
% &$\mathcal{T}_\mathcal{V}$ (Ours) & \textbf{93.60}  & \textbf{78.20} & \textbf{57.20} \\

\multirow{1}{*}{InternVL2-8B}
&$I$ & 91.12 & 75.40 & 55.64 & 74.05\\
\multirow{1}{*}{InternLM}
&$\mathcal{T}_\mathcal{V}$ (Ours) & \textbf{93.60} & \textbf{78.20} & \textbf{57.20} & \textbf{76.33}\\
\bottomrule
\end{tabular}
\vspace{-0.2cm}
\caption{\textbf{Additional VLMs for ARR on Small-set}. InternLM is `InternLM2-5-7b-chat'.}
\label{tab:rebuttal_more_vlms_smallset}
\vspace{-4pt}
\end{table}

\textbf{Comparison against more VLMs} In Tab.~\ref{tab:rebuttal_more_vlms_smallset}, we compare our method with two state-of-the-art VLMs: LLaVA 1.6 \cite{liu2023improvedllava} and InternVL2-8B \cite{Chen_2024_CVPR}. We utilized the language models from these models (InternLM \cite{cai2024internlm2} against InternVL2-8B and Vicuna-13B \cite{vicuna2023} against LLaVA 1.6) to retrieve correct action-reason statements based on descriptions generated by LLaVA 1.5 and LLaVA 1.6 respectively. The results show that our approach, using InternLM, outperforms InternVL2-8B, even when using LLaVA 1.5 verbalization, and Vicuna-13B when using LLaVA 1.6 verbalization, outperforms LLaVA 1.6.

\begin{table}[tp]
    \centering
    \scriptsize
    
    \begin{tabular}{l||ccc|ccc}
    \toprule
    \multirow{2}{*}{\textbf{Classifier}} & \multicolumn{3}{c|}{\textbf{\textit{Precision@k}}} & \multicolumn{3}{c}{\textbf{\textit{Top-k Acc}}} \\

    & \textbf{k=1} & \textbf{k=2} & \textbf{k=3} &  \textbf{k=1} & \textbf{k=2} & \textbf{k=3} \\
    \hline
    \hline
    CLIP($I$) & 61.04 &  33.86 &  22.66 &  23.72 &  44.61 &  61.04   \\
    % CLIP($I+ \hat{s}_{IN}$)  &  42.47 &  22.47 &  14.98 &  13.44 &  28.51 &  42.47   \\
    CLIP ($I + T$) & 46.15 &  24.36 &  16.24 &  15.15 &  31.25 &  46.15  \\
    % CLIP($I+T+ \hat{s}_{IN}$) & 00.00 &  56.51 &  30.39 & 20.26 &  19.35 &39.81  \\
    \hline
    CLIP ($I + T + V$) (Ours)&   70.46 &  39.17 &  26.11 &  29.79 &  53.08 &  70.46  \\
    % CLIP ($I + T + V+ \hat{s}_{IN}$)&   00.00 &  00.00 &  00.00 &  00.00 &  00.00 &  00.00  \\
    CLIP ($I + T + V + IN + UH$)  (Ours) &   \textbf{72.35} &  \textbf{41.05} &  \textbf{27.40} &  \textbf{32.11} &  \textbf{54.37} &  \textbf{72.35}  \\
    CLIP ($I + \mathcal{T}_\mathcal{V}$)  (Ours) &   63.53 &  34.25 &  22.83 &  24.14 &  45.38 & 63.53  \\
    % CLIP ($I + \mathcal{T}_\mathcal{V}$ + $\hat{s}_{IN}$)  &   00.00 &  00.00 &  00.00 &  00.00 &  00.00 &  00.00  \\

    \bottomrule
    \end{tabular}
    \caption{\textbf{Evaluation of CLIP-based models on Full-set}. \textbf{Bolded} numbers indicate the best performing model.}
    \label{tab:clip_ablation}
\end{table}

%  final table
\setlength{\tabcolsep}{0.5pt}
\begin{table}[!tp]\centering

\scriptsize
\begin{tabular}{lc||c|c|c||c|c||c||c}\toprule

&& \multicolumn{6}{c||}{\textbf{\textit{Multi}}} & \textbf{\textit{Single}}\\

\textbf{Classifier}&\textbf{Verb.}& \multicolumn{3}{c|}{\textbf{\textit{Precision@k}}} & \multicolumn{2}{c||}{\textbf{\textit{Top-k Acc}}}& \multirow{2}{*}{\cellcolor{LGray}} & \\
&&\textbf{k=1} &\textbf{k=2} &\textbf{k=3} &\textbf{k=1} &\textbf{k=2}& \cellcolor{LGray}{\textbf{Avg}}& \textbf{Acc} \\
\hline
\hline
% CLIP & $I$ & 64.52 &34.48 & 22.98 & 20.97& 47.18 &64.52 & 20.97 \\
% CLIP (KAFA) &$I +T$& 47.18 & 25.40 & 16.94 & 15.73 & 32.26  & 47.18 & 15.73 \\
% \hline
% LLaVA &$I$ &\textbf{59.67} &\textbf{38.27} &26.06 &32.92 & 48.14&\textbf{59.67} &26.00 \\
% % BLIP-2 &$I$ &Failed &Failed &Failed &Failed &Failed &Failed &37.38 \\
% % MiniGPT4 &$I$ & & & & & & & \\
% GPT-4V &$I$ &56.40 &38.00 &\textbf{26.80} &22.00 & \textbf{49.60} &56.40 &\textbf{90.37} \\

% \hline
% \multirow{2}{*}{LLaVA} 
% &$-$ &- & - & - & - &-  & \cellcolor{LGray} - &-\\
% &$CoT$ & 67.32 & 49.03 & 29.23 & 30.32 & 53.68  & \cellcolor{LGray}  - & -\\

% \hline
% \hline

% \multirow{2}{*}{\textcolor{blue}{LLaVA}} & $I$ & 59.67 & 38.27 & 26.06 & 32.92 & 48.14 & 41.01 & - \\
% & $I + CoT$ & 66.53 & 42.24 & 28.29 & 33.06 & 46.49 & 43.32 & - \\
% \hline
% \hline
\multirow{3}{*}{Vicuna} 
&$V + T$ &64.11 & 41.53 & 27.69 & 24.19 &45.56  & \cellcolor{LGray} 34.62 &44.35\\
&$IN$ & 64.92 & 43.55 & 29.17 & \textbf{24.60} & 43.55  & \cellcolor{LGray}  41.16 & 45.56\\
&$UH$ & 60.89 & 38.71&  25.94 & 20.97 & 40.32  & \cellcolor{LGray}  37.37 & 37.90\\
\hline
& $V + T + IN + UH$ & 69.35 & 45.33 & 30.88 & 23.80 & 45.21  & \cellcolor{LGray}  42.91& 48.39 \\

% small: &$V + T + IN + UH$ & 71.37 & \textbf{47.58} & \textbf{32.12} & \textbf{27.42} & \textbf{48.39 }& 71.37 & 48.39\\

\multirow{2}{*}{Ours (Vicuna)} &$\mathcal{T}_\mathcal{V} \setminus UH$ & 69.35 &  44.35 & 29.57 & 22.98 & 45.56  & \cellcolor{LGray} 42.36 & 48.39 \\

&$\mathcal{T}_\mathcal{V}$ &71.37 & \textbf{46.77} & 31.45 & 23.39 & 45.16  & \cellcolor{LGray} 43.63& 46.37\\

    &$\mathcal{T}_\mathcal{V}+\hat{s}_{IN}$ & \textbf{71.77} &\textbf{ 46.77} & \textbf{31.59} & 23.79 & \textbf{46.77}   &\cellcolor{LGray}  \textbf{44.14} & \textbf{48.79}\\
 
 % &$\mathcal{T}_\mathcal{V} + \hat{s}_{IN}$ &68.32 & 44.52 &30.25&22.95& 43.24 & 68.32 \\

% \multirow{2}{*}{Ours(Vicuna) with GPT4 verb} &$\mathcal{T}_\mathcal{V}$ &\textbf{72.18} & \textbf{47.18} &31.99 &28.23 &48.39 & \textbf{72.18} & \textbf{48.39 }\\
% &$\mathcal{T}_\mathcal{V} + \hat{s}_{UH}$ &  71.37 & \textbf{47.18} &\textbf{ 32.12 }& \textbf{29.44} & \textbf{48.79} & 71.37 &  47.98\\

% &$\mathcal{T}_\mathcal{V} + \hat{s}_{\mathcal{T}_\mathcal{V}}$ & 71.77 & 46.77 & 31.59 & 27.82 & \textbf{48.39} & 71.77 & \textbf{48.39} \\

\hline
\hline

\multirow{4}{*}{GPT-3.5} &$V + T$ &85.43 &59.51 &40.62 &46.56 &68.02  
 &\cellcolor{LGray} 60.02& 72.46 \\ %&91.50 \\
&$IN$ & 89.07& 64.78&  45.48 & \textbf{65.99} & 79.76  & \cellcolor{LGray} 69.01& 77.41 
 \\  %& \textbf{91.90} \\
&$UH$ & 84.62& 58.91& 40.89 & 52.63 &70.45   & \cellcolor{LGray} 61.10& 76.61 \\ %&&89.88 \\

\hline

&$V+T+IN+UH$ &90.32 &64.92 &45.43 &48.39 &74.60 & \cellcolor{LGray} 64.73 &  74.39 \\ %&90.32 \\

\multirow{2}{*}{Ours (GPT-3.5)} &$\mathcal{T}_\mathcal{V} \setminus UH$ & 91.09 & 66.81 & 46.55 & 62.75 & 78.94  & \cellcolor{LGray} 69.23 & 78.54 \\ %&78.86\\ 

&$\mathcal{T}_\mathcal{V}$ &\textbf{91.90} & \textbf{67.61} & \textbf{46.96} & 63.15 & 79.75  & \cellcolor{LGray} 69.87 & \textbf{78.94 }\\ %&& 78.13\\
 &$\mathcal{T}_\mathcal{V} + \hat{s}_{IN}$ & \textbf{91.90} & 67.20 & 46.69 & \textbf{65.99} & \textbf{81.78}  & \cellcolor{LGray} \textbf{70.71} &  72.87 \\ %74.38 \\ %&\\

% \multirow{3}{*}{Ours(GPT 3.5) + GPT4 verb}&$\mathcal{T}_\mathcal{V}$ &90.28 &\textbf{ 65.38} & \textbf{45.61} &\textbf{ 68.42 }& 72.87 &90.28 & \textbf{92.31} \\
% &$\mathcal{T}_\mathcal{V} + \hat{s}_{UH}$ &89.43 &63.82 &44.31 &64.63 & 70.37 &89.43 &91.90 \\

% &$\mathcal{T}_\mathcal{V} + \hat{s}_{\mathcal{T}_\mathcal{V}}$& \textbf{90.65} &64.23 &44.72 &65.85 &71.66 &\textbf{ 90.65} &91.90 \\

\hline
\hline

\multirow{4}{*}{GPT-4} &$V + T $&92.71 &84.62 &72.47 &84.55 & 89.52  & \cellcolor{LGray} 84.77 &95.55 \\
&$IN$ & 89.92& 78.23& 64.65&77.42 &85.08  & \cellcolor{LGray} 79.06 & 93.88 \\
&$UH$ & 80.41& 63.67 & 50.48&62.04 &72.65  & \cellcolor{LGray} 64.85 &95.08 \\

\hline

&$V+T+IN+UH$ &96.37 &86.49 &72.45 &72.18 &89.11 & \cellcolor{LGray} 83.32 &96.37 \\

\multirow{3}{*}{Ours (GPT-4)} &$ \mathcal{T}_{\mathcal{V}} \setminus UH$ & 94.34 & 85.63& 73.42 & 84.62 & 90.28  & \cellcolor{LGray} 85.66  & 88.21\\

&$\mathcal{T}_\mathcal{V}$ &96.77 &87.30 &\textbf{74.60} &84.96 &91.46 & \cellcolor{LGray} 87.01 & \textbf{96.77} \\

% &$\mathcal{T}_\mathcal{V} + \hat{s}_{UH}$&96.76 & \textbf{88.06 }& 74.09 & \textbf{86.64} & 90.69 & 96.76 &96.36 \\

&$\mathcal{T}_\mathcal{V} + \hat{s}_{IN}$ &96.77 &\textbf{87.77} &73.65 &\textbf{87.09} & \textbf{91.54} & \cellcolor{LGray} \textbf{87.36} & 96.36 \\

&$\mathcal{T}_\mathcal{V} + \hat{s}_{\mathcal{T}_\mathcal{V}}$ &\textbf{97.17 }&86.99 &73.55 &85.02 &91.50  & \cellcolor{LGray} 86.85 &96.76 \\

\bottomrule
\end{tabular}
\caption{\textbf{ARR on Small-set}. Best result per LLM/column is bolded. `Multi' means we ask the LLM for multiple outputs, `Single' for one.
}
\label{tab:tab_arr_small}
\end{table}

\textbf{CLIP ablation.}
Table~\ref{tab:clip_ablation} demonstrates different verbalization strategies impact on CLIP zero-shot model. We observe that in contrast to Table~\ref{tab:tab_arr_small} and Table 3 in the main paper, where the best results are mostly based on $\mathcal{T}_\mathcal{V}$, simple concatenation (i.e., $U+T+IN+UH$) achieves the best performance on CLIP.  This can be due to the more fine-grained (even noisy) information in $T+V+IN+UH$. Therefore, CLIP that has shown to have bag-of-words behavior \cite{yuksekgonul2023when} performs better when more information, such as object names, relations, etc., are explicitly noted. However, our proposed LLM-based approaches have more reasoning capabilities. Thus, a less noisy and more unified description in $\mathcal{T}_{\mathcal{V}}$ is a more suitable verbalization strategy. 

\textbf{Hard Negative Validation.} To ensure the quality of the generated hard negatives using GPT-4, we sampled 100 images and had 5 human annotators classify each option (options constitute both ground-truth action-reason statements and the generated hard negative options using our proposed method) as negative or positive. Here, `positive' indicates a correct action-reason statement for the corresponding image. Our observations revealed that 99.28\% were marked as true negatives by the annotators. Specifically, out of 1669 hard negative action-reason statements generated by the LLM, only 12 statements were identified as correct (i.e., positive), while 1657 were marked as incorrect (i.e., negative) action-reason statements for the images. This underscores the effectiveness of our method in generating valid, high-quality, semantically hard negatives for the action-reason retrieval task. 

\begin{figure}[t]
    \centering
     \begin{subfigure}[b]{0.48\textwidth}
        \includegraphics[width=\textwidth]{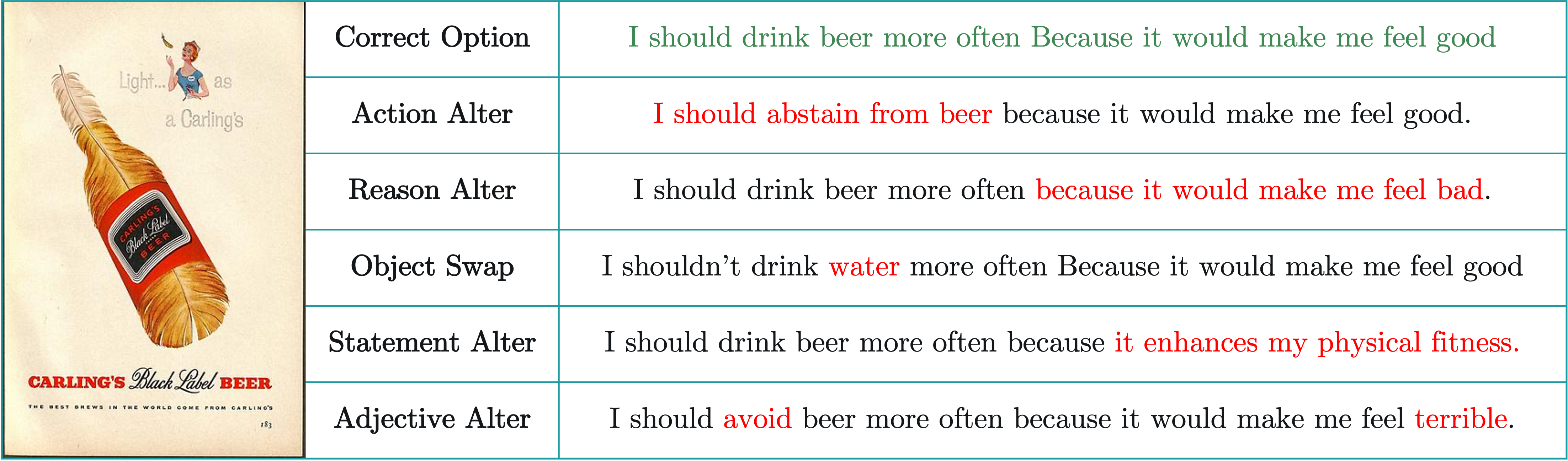}
        \label{fig:HN_examples_sub1}
    \end{subfigure}
    \begin{subfigure}[b]{0.48\textwidth}
        \includegraphics[width=\textwidth]{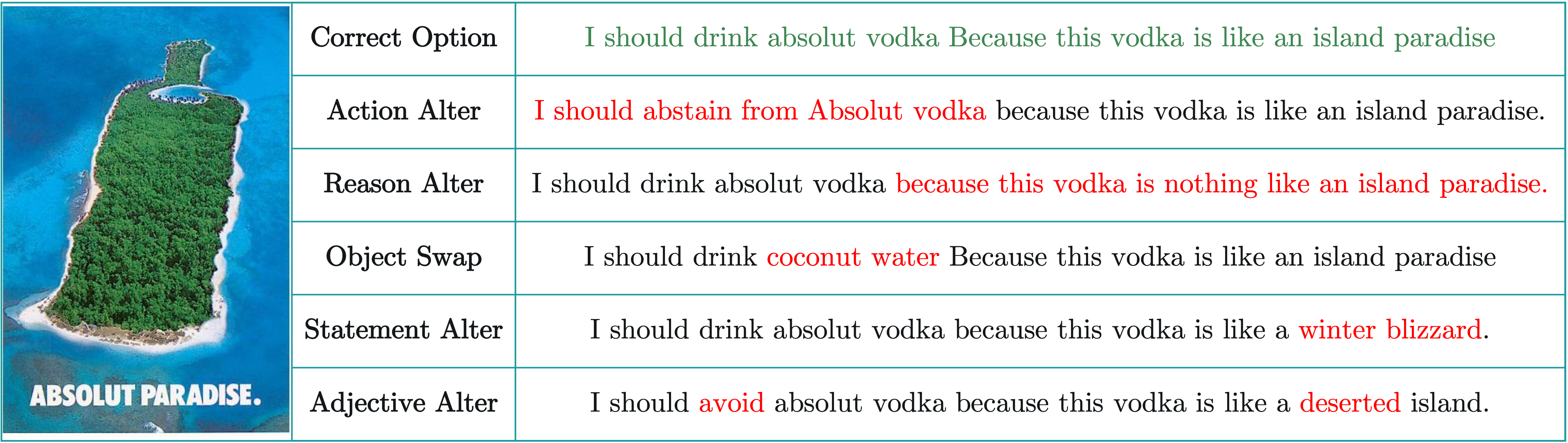}
        \label{fig:HN_examples_sub2}
    \end{subfigure}
    \begin{subfigure}[b]{0.48\textwidth}
        \includegraphics[width=\textwidth]{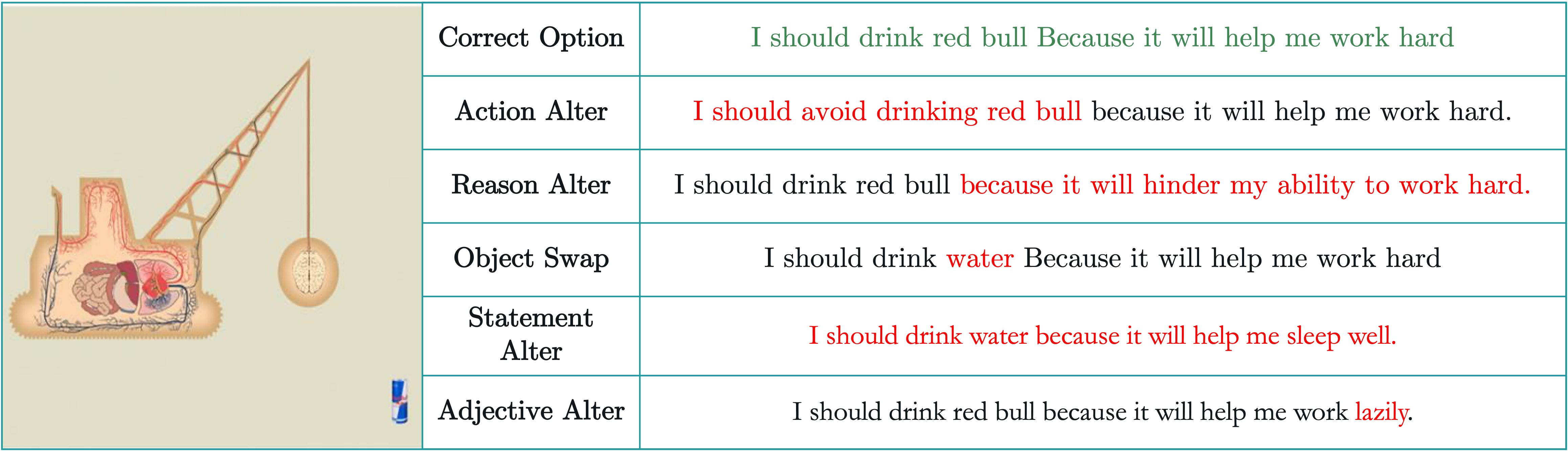}
        \label{fig:HN_examples_sub3}
    \end{subfigure}
    % \begin{subfigure}[b]{0.98\textwidth}
    %     \includegraphics[width=\textwidth]{images/supplement/HNexample4.png}
    %     \label{fig:HN_examples_sub4}
    % \end{subfigure}
    % \begin{subfigure}[b]{0.98\textwidth}
    %     \includegraphics[width=\textwidth]{images/supplement/HNexample4.png}
    %     \label{fig:sub3}
    % \end{subfigure}
    \caption{For each correct action-reason statement, we construct 5 different types of hard negatives: (1) Action Alter, (2) Reason Alter, (3) Object Swap, (4) Statement Alter, and (5) Adjective Alter. \textcolor{Green}{Green} denotes correct action-reason statements. \textcolor{red}{Red} indicates generated wrong phrases/statements.}
    \label{fig:HN_examples}
\end{figure}

 % As shown in the 4th example, some of the positive ground truths are not accurate. For example, in this example, the correct option is \textcolor{red}{`I should not buy Red Bull Because it destroys my inside.'} which is not an accurate message for an image advertising Red Bull.
\begin{figure*}[!htp]
        \includegraphics[width=\linewidth]{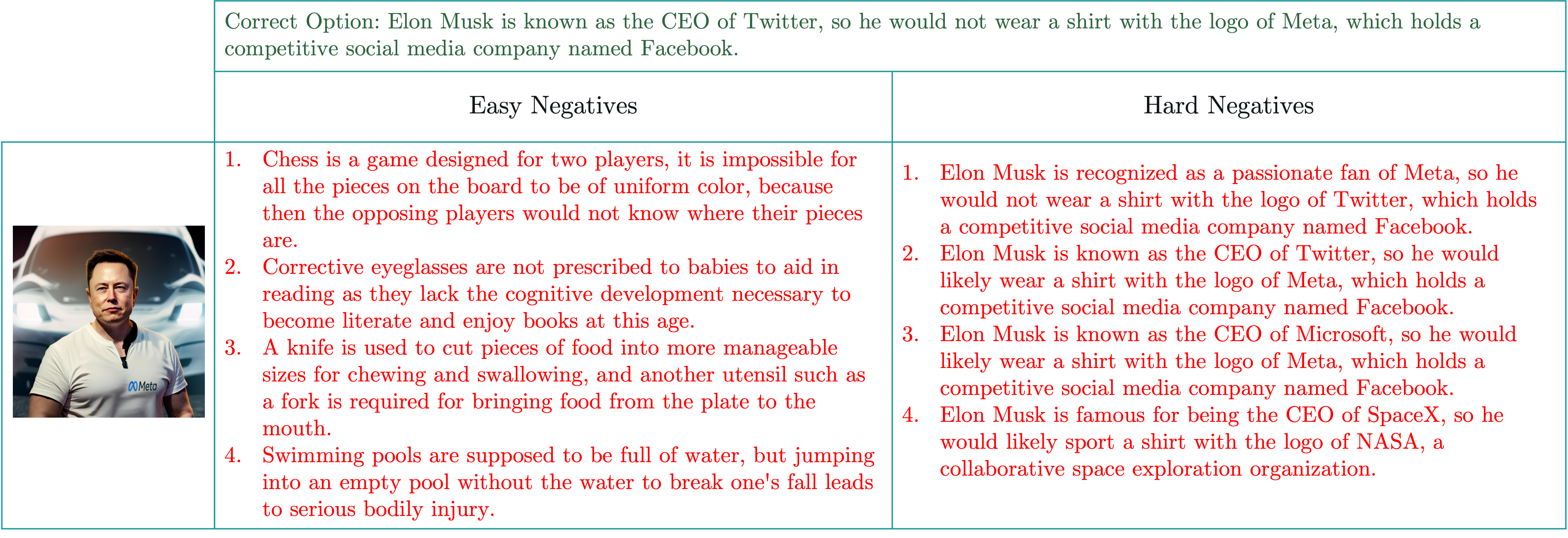}
        % \caption{}
        % \label{fig:intro_a}
    
    \caption{An example of Explanation task in WHOOPS! dataset \cite{whoops!} with easy and hard negative options. \textcolor{Green}{Green} shows correct option and \textcolor{red}{Red} shows incorrect options. }
    \label{fig:whoops_example}

\end{figure*}

Fig. \ref{fig:HN_examples} shows different types of hard negatives generated by GPT-4 for three images.

\textbf{BLIP-2 Failure.}
While we reported the performance of BLIP-2 \cite{li2023blip} for the AOR task, it was not effective for other tasks. For instance, BLIP-2 failed to follow instructions and produce reasonable output for multi-option/multi-label tasks like multi-ARR and MAC. For example, in the multi-ARR task, BLIP-2 erroneously identified all provided options as correct action-reason statements when only three correct statements were required. This limitation could be due to the lack of instruction tuning in the pre-training phase of the BLIP-2 model compared to more recent models such as LLaVA \cite{liu2023improvedllava}. Consequently, we explored InstructBLIP, an instruction-tuned version of the BLIP-2 model.

\textbf{Effectiveness of each component in atypicality-aware verbalization.} To further evaluate the effectiveness of different steps in atypicality-aware verbalization on the performance of different LLMs on ARR tasks, we repeated the experiments on the small set. We used Vicuna, GPT-3.5, and GPT-4 as the LLMs. As observed in Table~\ref{tab:tab_arr_small} $\mathcal{T}_\mathcal{V} + \hat{S}_{IN}$ verbalization performs better, with all the LLMs. $V + T + IN + UH$ includes the atypicality; however, LLaVA generated descriptions might be noisy. Combining them and denoising the combination by an LLM improve the performance. Inspired by ASR task, we detect the atypicality statement for the image using $IN$ description. The results in Table~\ref{tab:tab_arr_small} shows directly adding the detected atypicality statement to the verbalization, rather than keeping it implicit, further improves performance.

\textbf{Generalization to typical images}
\begin{table}
\centering

%is based on $UH$. LLaVA + GPT-4 verbalization. 
%, $\hat{s}_{UH}$: atypc prediction based on UH}
% \scriptsize
\begin{tabular}{lc||c|c|c}
\toprule
 \textbf{Classifier} & \textbf{Verb}. & \multicolumn{3}{c}{\textit{\textbf{Precision@k}}} \\

&&\textbf{k=1} & \textbf{k=2} &\textbf{k=3}\\
\hline
\hline
 LLaVA \cite{liu2023improvedllava}& $I$& 66.4 & 42.2 & 28.3 \\
 \hline
\multirow{1}{*}{Vicuna~\cite{vicuna2023}} &$\mathcal{T}_\mathcal{V}$ (Ours) & \textbf{71.2} & \textbf{48.6}  & \textbf{33.2} \\
\bottomrule
\end{tabular}
\caption{\textbf{ARR on Typical images}}
\label{tab:typical_multi_selection_action_reason}

\end{table} 
PittAd dataset \cite{hussain2017automatic} includes both typical and atypical ad images. The focus of the experiments in the main paper is on the atypical images in the dataset. However, to evaluate the generalization of the proposed atypicality-aware verbalization method to images without atypicality, we used the typical images in the dataset. Results in Table~\ref{tab:typical_multi_selection_action_reason} show that even in images without atypicality, our atypicality-aware verbalization outperforms LLaVA, demonstrating its generalizability.

\subsection{Generalization beyond Ads (WHOOPS!)}

WHOOPS!~\cite{whoops!} generates common sense-defying images by placing normal objects in an unusual context. Unlike persuasive ads, WHOOPS! doesn't include atypical objects, and its unusualness isn't designed to convey specific messages. Hence it does not need the further reasoning ability required in ads to connect the unusualness to the final message of the image. Despite these differences, we use WHOOPS! as the closest existing benchmark to test our atypicality-aware verbalization method beyond ads. Specifically, we focus on the Explanation task, which involves identifying an explanation for why an image is unusual.

Initially, we used 15 random explanations as negative options, but this is inadequate to effectively evaluate the reasoning ability of the models. These negatives may be unrelated to the image's scene/content, contain objects absent in the image, or describe irrelevant actions. As a result, models could easily eliminate these options using basic image understanding, such as object recognition. For example, in Fig.~\ref{fig:whoops_example} (left), a model could simply rule out all options due to mentioning `chess,' `babies,' `knife,' or `swimming pool' - objects clearly not in the image. Such easy negatives fail to effectively evaluate models' reasoning and deeper image understanding capabilities.

To address this limitation, we employed GPT-4 to generate more challenging negative options by (1) \textit{Random Options}: randomly chosen from the explanation of other images; (2) \textit{Alter Verb}: replacing a verb in the correct explanation with another verb and changes the meaning of the sentence; (3) \textit{Alter Object}: replacing an effective object in the correct explanation with an object visually similar to the original object; (4) \textit{Alter Adjective}: replacing an adjective in the correct explanation or add an adjective that changes the sentence semantically; and (5) \textit{Alter Causal}: changing the second half of the correct explanation while keeping the first half unchanged. Unlike easy negatives, these options (right column in Fig.~\ref{fig:whoops_example}) are closely related to the image content, making simple object recognition insufficient. Instead, these hard negatives demand deeper reasoning and more nuanced analysis.

Table~\ref{tab:whoops} shows that LLaVA (i.e., LLaVA outperforms Vicuna with $\mathcal{T}_\mathcal{V}$ verbalization with easy negatives. In contrast, Vicuna($\mathcal{T}_\mathcal{V}$) has better performance on the Explanation task with hard negative options. This demonstrates that our proposed atypicality-aware verbalization method generalizes on \textbf{unusual images beyond ads}, especially when metaphorical reasoning is required to fully interpret the image.

%  final table
\setlength{\tabcolsep}{2pt}
\begin{table}[!tp]\centering
% \footnotesize
\begin{tabular}{lc||c||c}\toprule

\textbf{Classifier}&\textbf{Verb.}& {\textbf{Explanation Hard}} & \textbf{Explanation Easy} \\
\hline
\hline
\multirow{1}{*}{LLaVA} & \_ &18.8 & \textbf{88.0} \\

\multirow{1}{*}{Vicuna}

&$\mathcal{T}_\mathcal{V}$ &\textbf{20.4} & 65.4 \\

\bottomrule
\end{tabular}
\caption{\textbf{Explanation results on Whoops dataset}. Evaluation metric for Explanation is accuracy. Explanation Easy indicates Explanation task with hard negative options generated by GPT-4 and Explanation Easy indicates Explanation task with negative options randomly chosen from the explanation of other images.
}
\label{tab:whoops}
\end{table}

\subsection{Prompts}
\label{supp:prompts}

    % In the third row, our method retrieves 2 out of 3 correct statements, compared to LLaVA's 1 out of 3. 
    % However, this negative, `I should buy Red Bull because it nourishes my insides,' might be considered a plausible action-reason statement for the image. In fact, This false negative, generated by our Semantically Hard Negative generator, contrasts with the statement, `I should not buy Red Bull because it destroys my insides,' which is erroneously marked as the ground truth in the dataset.
    
% \begin{figure}
%     \centering
    
%     \caption{Caption}
%     \label{fig:enter-label}
% \end{figure}

Throughout our experimentation, we explored various prompt strategies for each LLM (i.e., Vicuna and GPT models). We utilized a fixed prompt for each task that achieved the best performance for the respective LLM, ensuring adherence to the instructions and output format. It's important to note that all methods were implemented using the same prompt for a given LLM to ensure correctness and fair evaluation. 

\textbf{Verbalization prompts.}
Prompts utilized to verbalize the image and obtain `list of top-5 objects' ($V$), `text-scene' ($T$), `image description' ($IN$), and `unusualness' ($UH$) are depicted in Listing~\ref{list:prompt_V_llava}, Listing~\ref{list:prompt_T_llava}, Listing~\ref{list:prompt_IN_llava}, Listing~\ref{list:prompt_UH_llava}, respectively. GPT4-V prompts use the same question without LLaVA's specific prompt format (i.e., `USER:$<$image$>$' and `ASSISTANT:'). Finally, Listing~\ref{list:prompt_combine_gpt4} illustrates the prompt for combining the LLaVA/GPT-4V verbalization to obtain $\mathcal{T}_\mathcal{V}$ for both GPT-4 and Vicuna.

\textbf{Atypicality Understanding prompts.}
Listing~\ref{list:prompt_mac_gpt} and Listing~\ref{list:prompt_mac_vicuna} showcase the Multi-label classification (MAC) prompt templates for GPT and Vicuna models, respectively. Listing~\ref{list:prompt_asr_gpt} and Listing~\ref{list:prompt_asr_vicuna} are Atypical Statement Retrieval (ASR) prompt templates for GPT and Vicuna, respectively. See Listing~\ref{list:prompt_aor-gpt_llava} for GPT and LLaVA, Listing~\ref{list:prompt_aor-minigpt4} for MiniGPT4, and Listing~\ref{list:prompt_aor-blip2_instructblip} for BLIP2 and InstructBLIP, for examples of the prompts used in the Atypicality Object Recognition (AOR) task.

\textbf{Action-Reason Retrieval prompts.}
Listing~\ref{list:prompt_arr_single_gpt} and Listing~\ref{list:prompt_arr_multi_gpt} exhibits prompt templates for GPT-based language models for single-ARR and multi-ARR tasks. The corresponding prompts for the Vicuna language model can be found in Listing~\ref{list:prompt_arr_single_vicuna} and Listing~\ref{list:prompt_arr_multi_vicuna} for the single and multi-tasks, respectively. 

\centering
\begin{lstlisting}[caption={LLaVA's prompt for list of top-5 objects $V$},label={list:prompt_V_llava}]
(*@\textbf{USER}@*): 
 (*@$<$image$>$@*)
What are the non-textual objects visible in this image? Carefully output AT MOST top 
5 objects. If there are more than 5 objects, output major/important objects
according to the image. Words/Texts are not considered as objects. Separate 
with a comma.
(*@\textbf{ASSISTANT}@*):
\end{lstlisting}

\begin{lstlisting}[caption={LLaVA's prompt for text-scene $T$},label={list:prompt_T_llava}]
(*@\textbf{USER}@*): 
(*@$<$image$>$@*)
You are an OCR expert. What are the text messages in the image? If there are no text
messages on the image, return only `NO TXT'
(*@\textbf{ASSISTANT}@*):
\end{lstlisting}

\begin{lstlisting}[caption={LLaVA's Prompt for image description $IN$},label={list:prompt_IN_llava}]
(*@\textbf{USER}@*):
(*@$<$image$>$@*)
Describe the image in detail.
(*@\textbf{ASSISTANT}@*):
\end{lstlisting}

\begin{lstlisting}[caption={LLaVA's prompt for unusualness $UH$},label={list:prompt_UH_llava}]
(*@\textbf{USER}@*): 
(*@$<$image$>$@*)
What is unusual about this image?
(*@\textbf{ASSISTANT}@*):
\end{lstlisting}

\begin{lstlisting}[caption={\textbf{GPT-4 and Vicuna Prompt template for combining LLaVA's/GPT-4V verbalizations to generate $\mathcal{T}_\mathcal{V}$}. \textcolor{highlight-blue}{\{Blue\}} denotes elements added dynamically.},label={list:prompt_combine_gpt4}]
A chat between a curious human and an artificial intelligence assistant. The assistant gives helpful, detailed, and polite answers to the human's questions about an image. Analyze each assistant's response carefully, then combine the information, and summarize the combined information in a way that is useful for further question/answering tasks. Your answer must be a high-quality summary of the information.

(*@\textbf{Input Question-Anwers:}@*)

Question: 
    What are the non-textual objects visible in this image? Carefully output AT MOST top 5 objects. If there are more than 5 objects, output major/important objects according to the image. Words/Texts are not considered as objects. Separate with comma.
Answer:
      (*@\textcolor{highlight-blue}{\{$V$ (List of top-5 objects)\}}@*)
Question:
    You are an OCR expert. What are the text messages in the image?
Answer:
    (*@\textcolor{highlight-blue}{\{$T$ (List of scene-tests)\}}@*)
Question:
    Describe the image in detail.
Answer:
    (*@\textcolor{highlight-blue}{\{$IN$ (ImageNarrator)\}}@*)
Question:
    What is unusual about this image?
Answer:
    (*@\textcolor{highlight-blue}{\{$UH$ (UnusualHighlighter)\}}@*)
Carefully combine the above information given by the assistant.\end{lstlisting}

\begin{lstlisting}[caption={\textbf{GPT prompt template for Mac}. \textcolor{highlight-blue}{\{Blue\}} denotes elements added dynamically.},label={list:prompt_mac_gpt}]
Consider the following atypicality definition:
       (*@\textcolor{highlight-blue}{\{$\mathcal{D}_\mathcal{A}\ $atypicality definition\}}@*)
Use the above definitions to help the user in classifying atypicalities in the images.
(*@\textbf{Question}@*):
You are a highly intelligent and accurate image atypicality multi-label classification system. You take an Image Description as input and classify that into at most 4 appropriate atypicality Categories from the given category list:
    (1) TR1
    (2) TR2
    (3) OIO
    (4) OR
You should select multiple atypicality categories ONLY if multiple atypicalities are present in the image.
If none of the atypicality categories exist, one of the predicted labels has to be "NA."
Your output format is only {{ output_format|default("[{'1': 1st level Atypicality Category, '2': 2nd level Atypicality Category,...}]") }} form, no other form.
(*@\textbf{Image Description}@*):
        (*@\textcolor{highlight-blue}{\{image-description (e.g., $UH$)\}}@*)
\end{lstlisting}

\begin{lstlisting}[caption={\textbf{Vicuna prompt template for MAC}. \textcolor{highlight-blue}{\{Blue\}} denotes elements added dynamically.},label={list:prompt_mac_vicuna}]
(*@\textbf{USER}@*): You are a highly intelligent multi-label classification system. You will be given an Image Description and a Question. Answer the question based on the Image Description:
(*@\textbf{Image Description:}@*)
    (*@\textcolor{highlight-blue}{\{description (e.g., $UH$)\}}@*)
(*@\textbf{Question:}@*)
According to the Image Description and the atypicality definitions below, detect the atypicality categories:
    (*@\textcolor{highlight-blue}{\{$\mathcal{D}_\mathcal{A}\ $atypicality definition\}}@*)
You should select multiple atypicality categories ONLY if multiple atypicalities are present in the image.
If none of the atypicality categories exist, one of the predicted labels has to be "NA".
You must choose the detected atypicalities from (OIO, TR1, TR2, and OR) and the acceptable output format is only {{ output_format|default("[{'1': 1st level Atypicality Category, '2': 2nd level Atypicality Category,...}]") }} form, no other form. Do NOT output any extra information or explanation.
(*@\textbf{ASSISTANT}@*):
\end{lstlisting}

\begin{lstlisting}[label={list:prompt_asr_gpt}, caption={\textbf{GPT prompt template for Atypical Statement Retrieval (ASR)}. \textcolor{highlight-blue}{\{Blue\}} denotes elements added dynamically.}]
(*@\textbf{Context}@*): (*@\textcolor{highlight-blue}{\{description (e.g., IN)\}}@*)
(*@\textbf{Question}@*): Based on the context return the index of best statement among the options to describe the relation among the objects in the context.
If the context does not provide helpful information, choose the best option from your side.
(*@\textbf{Options}@*): (*@\textcolor{highlight-blue}{\{list of generated correct and incorrect atypicality statements\}}@*)
Your output format is only (*@\textcolor{black}{Answer: \$\{index of correct statement\}}@*) form, no other form. None of the above is not allowed. Even with not enough context and information, you must choose one of the options based on an unusual part of the context.
\end{lstlisting}

\begin{lstlisting}[label={list:prompt_asr_vicuna}, caption={\textbf{Vicuna prompt template for Atypical Statement Retrieval (ASR)}. \textcolor{highlight-blue}{\{Blue\}} denotes elements added dynamically.}]
(*@\textbf{USER}@*):
(*@\textbf{Context}@*): (*@\textcolor{highlight-blue}{\{IN description\}}@*)
(*@\textbf{Question}@*): Based on the context return the index of best statement among the options to describe the relation among the objects in the context.
If the context does not provide helpful information, choose the best option.
(*@\textbf{Options}@*):  (*@\textcolor{highlight-blue}{\{list of generated correct and incorrect atypicality statements\}}@*)
Your output format is only (*@\textcolor{black}{Answer: \$\{index of correct statement\}}@*) form, no other form. None of the above is not allowed. Even with not enough context and information, you must choose one of the options based on an unusual part of the context.
(*@\textbf{ASSISTANT}@*):
\end{lstlisting}

\begin{lstlisting}[caption={\textbf{GPT and LLaVA prompt template for Atypical Object Recognition (AOR)}. \textcolor{highlight-blue}{\{Blue\}} denotes elements added dynamically based on the atypicality relation. Here, we show the TR1 atypicality relation as an example.},label={list:prompt_aor-gpt_llava}]
(*@\textbf{USER}@*):
(*@$<$image$>$@*)
A human has described this image as atypical. They have found it atypical because of: (*@\textcolor{highlight-blue}{Texture Replacement 1, with objects' texture borrowed from another object}@*).
More specifically, (*@\textcolor{highlight-blue}{The surface of $<$object1$>$ mimics the texture of $<$object2$>$, while retaining its original structure}@*).
Fill in your answers for (*@$<$object1$>$@*) and (*@$<$object2$>$@*). Make sure to include the angular brackets (*@$<$ and $>$@*).
An example output: (*@\textcolor{highlight-blue}{The surface of <eleven> mimics the texture of $<$meat$>$, while retaining its original structure}@*).
(*@\textbf{ASSISTANT}@*):
\end{lstlisting}

\begin{lstlisting}[caption={\textbf{MiniGPT4 prompt template for Atypical Object Recognition (AOR)}. \textcolor{highlight-blue}{\{Blue\}} denotes elements added dynamically based on the atypicality relation. Here, we show the TR1 atypicality relation.},label={list:prompt_aor-minigpt4}]
(*@\textbf{USER}@*):
(*@$<$image$>$@*)
A human has described this image as atypical. They have found it atypical because of: (*@\textcolor{highlight-blue}{Texture Replacement 1, with objects' texture borrowed from another object}@*).
More specifically, (*@\textcolor{highlight-blue}{The surface of $<$object1$>$ mimics the texture of $<$object2$>$, while retaining its original structure}@*).
Give short answers for what (*@$<$object1$>$@*) and (*@$<$object2$>$@*) are, in the format:
(*@$<$object1$>$@*): (*@$<$answer1$>$@*)
(*@$<$object2$>$@*): (*@$<$answer2$>$@*)
(*@\textbf{ASSISTANT}@*):\end{lstlisting}

\begin{lstlisting}[caption={\textbf{BLIP2 and InstructBLIP prompt template for Atypical Object Recognition (AOR)}. We use a multi-step prompt to generate the primary and secondary objects separately. \textcolor{highlight-blue}{{\{Blue\}}} denotes elements added dynamically based on the atypicality relation. Here, we show the TR1 atypicality relation.},label={list:prompt_aor-blip2_instructblip}]
<image>
A human has described this image as atypical. They have found it atypical because of: (*@\textcolor{highlight-blue}{Texture Replacement 1, with objects' texture borrowed from another object}@*).
More specifically, (*@\textcolor{highlight-blue}{The surface of $<$object1$>$ mimics the texture of $<$object2$>$, while retaining its original structure}@*).
Give short answers for what (*@$<$object1$>$@*) and (*@$<$object2$>$@*) are.
(*@$<$object1$>$@*): (*@\textbf{VLM prompted here}@*)
(*@$<$object2$>$@*): (*@\textbf{VLM prompted here}@*)\end{lstlisting}

\begin{lstlisting}[caption={\textbf{GPT prompt template for Action-Reason Retrieval (ARR) choosing single correct option}. \textcolor{highlight-blue}{\{Blue\}} denotes elements added dynamically.},label={list:prompt_arr_single_gpt}]
(*@\textbf{Context}@*): (*@\textcolor{highlight-blue}{ \{$\mathcal{T}_\mathcal{V}\ $description\} \{Atypicality statement\}}@*)
(*@\textbf{Question}@*): Based on the context return the index of the best statement among the options to interpret the described image.
Even without enough information return the index of the best  option among the options.
(*@\textbf{Options}@*): (*@\textcolor{highlight-blue}{\{list of correct and incorrect action-reason statements\}}@*)
Your output format is only (*@\textcolor{black}{Answer: \$\{index of correct statement\}}@*) form, no other form. 
None of the above is not allowed. Even without enough information choose the best interpretation. \end{lstlisting}

\begin{lstlisting}[caption={GPT prompt template for Action-Reason Retrieval (ARR) choosing all correct options. \textcolor{highlight-blue}{\{Blue\}} denotes elements added dynamically.},label={list:prompt_arr_multi_gpt}]
(*@\textbf{Context}@*): (*@\textcolor{highlight-blue}{\{$\mathcal{T}_\mathcal{V}\ $description\} \{Atypicality statement\}}@*)
(*@\textbf{Question}@*): Based on the context return the indices of the 3 best statements among the options to interpret the described image.
Separate the answers by comma and even without enough information return the indices of the 3 best  options among the options.
(*@\textbf{Question}@*): (*@\textcolor{highlight-blue}{\{list of correct and incorrect action-reason statements\}}@*)
Your output format is only (*@\textcolor{black}{Answer: \$\{indices of the 3 best statements\}}@*) form, no other form. 
None of the above is not allowed. Even without enough information choose the 3 best interpretations. \end{lstlisting}

\begin{lstlisting}[caption={\textbf{Vicuna prompt template for Action-Reason Retrieval (ARR) choosing single correct option}. \textcolor{highlight-blue}{\{Blue\}} denotes elements added dynamically.},label={list:prompt_arr_single_vicuna}]
(*@\textbf{USER}@*):
(*@\textbf{Context}@*): (*@\textcolor{highlight-blue}{\{$\mathcal{T}_\mathcal{V}\ $description\} \{Atypicality statement\}}@*)
(*@\textbf{Question}@*): Based on the context return the index of the best statement among the options to interpret the described image.
(*@\textbf{Options}@*): (*@\textcolor{highlight-blue}{\{list of correct and incorrect action-reason statements\}}@*)
None of the above is not allowed. Even without enough information, choose the best interpretations.
Your output format is only (*@\textcolor{black}{Answer: \$\{index of correct statement\}}@*) form, no other form. 
(*@\textbf{ASSISTANT}@*): \end{lstlisting}

\begin{lstlisting}[caption={\textbf{Vicuna prompt template for Action-Reason Retrieval (ARR) choosing all correct options}. \textcolor{highlight-blue}{\{Blue\}} denotes elements added dynamically.},label={list:prompt_arr_multi_vicuna}]
(*@\textbf{USER}@*):
(*@\textbf{Context}@*): (*@\textcolor{highlight-blue}{\{$\mathcal{T}_\mathcal{V}\ $description\} \{Atypicality statement\}}@*)
(*@\textbf{Question}@*): Based on the context, return the indices of the 3 best statements among the options to interpret the described image.
Separate the answers by comma, and even without enough information, return the indices of the 3 best options.
(*@\textbf{Options}@*): (*@\textcolor{highlight-blue}{\{list of correct and incorrect action-reason statements\}}@*)
None of the above is not allowed. Even without enough information, choose the 3 best interpretations.
Your output format is only (*@\textcolor{black}{Answer: \$\{indices of the 3 best statements\}}@*) form, no other form. 
(*@\textbf{ASSISTANT}@*): \end{lstlisting}

\begin{lstlisting}[caption={\textbf{Prompt for generating Action Alter hard negatives}.\textcolor{highlight-blue}{\{Blue\}} denotes elements added dynamically, based on the correct option.},label={prompt_action_hard}]
Generate one hard negative statement that semantically contradicts the action in the 
following correct statement. 
The hard negative should be plausible but must convey an opposite or entirely different
action, while the underlying reason remains unchanged. This requires reversing the
action's intent or suggesting a completely different concept that contrasts with
the original message, yet sounds coherent when paired with the same rationale.

(*@\textbf{Example:}@*)
    - Correct Statement: "I should get involved with artistic expression because dressing in style is a type of art."
    - Generated Hard Negative: "I should avoid artistic expression because dressing in style is a type of art."
In this example, "I should get involved with artistic expression" is the action, which is inverted to "I should avoid artistic expression" in the hard negative. 
The reason, "because dressing in style is a type of art," remains constant.

(*@\textbf{Correct Interpretation: }@*)(*@\textcolor{highlight-blue}{\{correct option\}}@*)

The hard negatives should closely mirror the vocabulary of the correct interpretation but must imply an opposite or distinctly different meaning. Only the hard negative statement is needed, without additional explanations.
\end{lstlisting}

\begin{lstlisting}[caption={\textbf{Prompt for generating Reason Alter hard negatives}. \textcolor{highlight-blue}{\{Blue\}} denotes elements added dynamically, based on the correct option.},label={prompt_reason_hard}]
Create a hard negative statement that presents semantically incorrect or opposite reasons compared to the provided correct statement while keeping the main action unchanged. These hard negatives should seem plausible at a glance but must convey a reason that contradicts the correct one. The intention is to maintain a surface-level similarity in wording with the original statement but to invert the underlying rationale.
                    
(*@\textbf{Example:}@*)
    - Correct Statement:``I should get involved with artistic expression because dressing in style is a type of art.''
    - Generated Hard Negative: ``I should get involved with artistic expression because dressing in style lacks artistic value.''

In this example, the action phrase ``I should get involved with artistic expression'' remains the same across both statements. The original reason, ``because dressing in style is a type of art'' is transformed to imply the opposite meaning, ``because dressing in style lacks artistic value,'' for the hard negative.

(*@\textbf{Guidelines:}@*)
1. Retain the action statement unchanged.
2. Invert the logic or reasoning of the correct statement to formulate the hard negative.
3. Ensure the hard negative retains similar wording to the original, but clearly communicates a contradictory reason.

(*@\textbf{Correct Interpretation: }@*)(*@\textcolor{highlight-blue}{\{correct option\}}@*)

Provide only the hard negative statement, ensuring it closely mirrors the correct interpretation in structure and vocabulary but distinctly opposes it in meaning.
                    
\end{lstlisting}

\begin{lstlisting}[caption={\textbf{Prompt for generating Statement Alter hard negatives}. \textcolor{highlight-blue}{\{Blue\}} denotes elements added dynamically, based on the correct option.},label={prompt_statement_hard}]
Generate a hard negative statement that is semantically unrelated and incorrect compared to a given correct statement. These hard negatives should be coherent statements on their own but must diverge completely in meaning from the original statement. The challenge is to craft a statement that, while maintaining superficial word similarity to the correct statement, introduces a concept or reasoning that is entirely irrelevant and incorrect.
                    
(*@\textbf{Example}@*):
    - Correct Statement: ``I should use 5-hour energy because it will keep me focused.''
    - Generated Hard Negative: ``I should use 5-hour stress drink because it promotes relaxation.''

(*@\textbf{Guidelines}@*):
1. Keep a superficial structural similarity to the correct statement in terms of wording.
2. Change the concept or reasoning to something totally irrelevant or even diametrically opposed to the original statement.
3. The hard negative should be plausible as a standalone statement but should not accurately reflect the logic or purpose of the correct interpretation.

(*@\textbf{Correct Interpretation:}@*) (*@\textcolor{highlight-blue}{\{correct option\}}@*)

Provide only the hard negative statement. It should closely mimic the correct statement in form but must diverge significantly in semantic content or meaning, introducing a totally different concept. 

\end{lstlisting}

\begin{lstlisting}[caption={\textbf{Prompt for generating Object Swap hard negatives}. \textcolor{highlight-blue}{\{Blue\}} denotes elements added dynamically, based on the correct option.},label={prompt_object_hard}]
Please generate a hard negative statement that has semantically incorrect (e.g., opposite) meaning to the one in the following correct statement by changing at least one object in the statement. Each hard negative should be a plausible option but must convey the incorrect meaning as the correct one. 

(*@\textbf{Example:}@*)
    - Correct statement: I should get involved with artistic expression Because dressing in style is a type of art 
    - Generated Incorrect statement: I should get involved with sports Because professional soccer is a type of sport
    

(*@\textbf{Correct Interpretation: }@*)(*@\textcolor{highlight-blue}{\{correct option\}}@*)

Ensure that the hard negatives maintain a degree of similarity to the correct interpretation in terms of words but imply incorrect meaning and include incorrect objects.
Only return the hard negative.
\end{lstlisting}

\begin{lstlisting}[caption={\textbf{Prompt for generating Adjective Alter hard negatives}. \textcolor{highlight-blue}{\{Blue\}} denotes elements added dynamically, based on the correct option.},label={prompt_adjective_hard}]
Given a correct statement, your task is to generate a hard negative statement. A hard negative statement should closely resemble the original statement in structure but convey a totally different meaning. This can be achieved by either changing an adjective to its antonym or by adding a qualifying adjective that totally changes the statement's sentiment. The goal is to create a plausible, yet semantically different version of the original statement. 
                    
(*@\textbf{The resulting hard negative should:}@*)
    - Only change or add an adjective
    - Keep the core structure of the original statement intact.
    - Alter the meaning to be totally different or even opposite by focusing on the modification of adjectives.
    - Ensure that the new statement is plausible and grammatically correct, but clearly wrong when compared to the original correct interpretation.

(*@\textbf{Example:}@*)
    - Correct Statement: ``I should use 5-hour energy because it will keep me focused.''
    - Hard Negative: ``I should use 5-hour energy because it will keep me sleepy.''

(*@\textbf{Correct Interpretation:}@*) (*@\textcolor{highlight-blue}{\{correct option\}}@*)

Please generate a hard negative based on the provided correct interpretation, focusing on the inversion of adjectives to create a totally different meaning.

\end{lstlisting}

\end{document}